\def\aospreprint{}
\documentclass{article}
\PassOptionsToPackage{hyphens}{url}

\usepackage[T1]{fontenc}
\usepackage{iclr2027_conference,times}

\usepackage{amsmath,amsfonts,bm}

\def\eqref#1{equation~\ref{#1}}

\def\1{\bm{1}}

\DeclareMathAlphabet{\mathsfit}{\encodingdefault}{\sfdefault}{m}{sl}
\SetMathAlphabet{\mathsfit}{bold}{\encodingdefault}{\sfdefault}{bx}{n}

\usepackage{booktabs}
\usepackage{array}
\usepackage{caption}
\usepackage{float}
\usepackage{wrapfig}
\usepackage{graphicx}
\usepackage{hyperref}
\usepackage{microtype}
\usepackage{url}
\usepackage{xcolor}

\definecolor{linknavy}{HTML}{123B6D}
\hypersetup{
  colorlinks=true,
  breaklinks=true,
  linkcolor=linknavy,
  citecolor=linknavy,
  urlcolor=linknavy,
  filecolor=linknavy,
  pdfborder={0 0 0},
  pdfborderstyle={/S/U/W 0}
}

\newcommand{\benchname}{AOS-Bench}
\expandafter\newcommand\csname benchname_explain\endcsname{%
The name \benchname{} reflects the vision of an agent serving as a ``natural-language operating system'' for managing users' computational workloads.%
}

\newcommand{\titlebench}{\texorpdfstring{{\normalfont\fontfamily{pzc}\fontshape{it}\fontsize{21}{24}\selectfont AOS-Bench}}{AOS-Bench}}
\ifdefined\aospreprint
\title{\parbox[c]{2.6em}{\includegraphics[width=\linewidth]{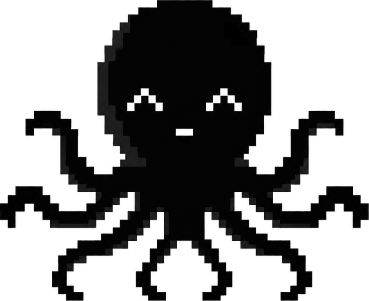}}\hspace{0.55em}\begin{tabular}[c]{@{}c@{}}\titlebench: Can LLM Agents\\ Manage Dynamic Workloads?\end{tabular}}
\author{Erle Zhu$^{1*}$\quad Genrui Zhang$^{2*}$\quad Keyang Hu$^{1}$\quad Mingyu Su$^{1}$\quad Hongning Wang$^{1\dagger}$ \\[2pt]
\normalfont $^{1}$Tsinghua University \qquad $^{2}$Columbia University
}
\iclrfinalcopy
\makeatletter
\def\@maketitle{\vbox{\hsize\textwidth\centering
{\LARGE\sc \@title\par}\vskip 0.22in
{\bf\@author\par}
\vskip 0.3in minus 0.1in}}
\makeatother
\newcommand{\aosfigonewidth}{0.9\linewidth}
\newcommand{\aoscontribspacing}{\setlength{\itemsep}{0pt}\setlength{\parsep}{0pt}}
\else
\newcommand{\aosfigonewidth}{\linewidth}
\newcommand{\aoscontribspacing}{}
\title{\titlebench: Can LLM Agents Manage\\ Dynamic Workloads?}
\author{Anonymous authors}
\fi

\begin{document}

\maketitle
\ificlrfinal
\lhead{Preprint}
{\renewcommand{\thefootnote}{\fnsymbol{footnote}}%
\footnotetext[1]{Equal contribution: \texttt{erlerzhu@gmail.com}, \texttt{gz2430@columbia.edu}.}%
\footnotetext[2]{Corresponding author: \texttt{hw-ai@tsinghua.edu.cn}.}}
\renewcommand{\thefootnote}{\fnsymbol{footnote}}\setcounter{footnote}{2}
\fi

\begin{abstract}
Coding agents are moving from handling one request at a time to serving as persistent assistants to which users hand a stream of work: a build, a data-processing job and a bug fix may all be pending on the same machine at once.
Such an agent must decide when to take on new tasks, how to delegate and overlap them, and how to stay within the machine's CPU and memory.
Existing benchmarks cannot measure this, because they evaluate each task in isolation.
We introduce \benchname{}, which evaluates an agent system as a workload manager: verified tasks arrive over time and must be delivered under shared CPU and memory limits, and the same tasks run one at a time serve as a per-system reference.

Across five model--harness pairs, isolated task success does not carry over evenly.
On heavy workloads every system loses 13 to 23 points.
On medium workloads, systems span 11 points in isolated success but 26 points in stream success, and two systems 0.6 points apart in isolation deliver 54\% and 68\% of the same tasks.
Execution-policy interventions trace much of this gap to task intake: serial dispatch leaves 73\% of heavy-workload tasks unclaimed.
Yet more elaborate management does not necessarily help.
A fixed concurrency of two comes within four points of the best agent-managed policy on both medium and heavy workloads, and under every policy median CPU use stays below 30\% of the provisioned cores while tasks wait.
These results identify workload management as a distinct capability for agents, and \benchname{} as a way to measure it.
\ifdefined\aospreprint
Code is available at \url{https://github.com/AOS-Bench/AOS-Bench}.
\fi
\end{abstract}

\section{Introduction}
\label{sec:introduction}

Advances in coding and computer-use agents let users delegate a growing range of computer tasks through natural language.
Benchmarks such as Terminal-Bench and OSWorld typically use individual tasks as the unit of evaluation~\citep{terminalbench,osworld}.
New requests may arrive before earlier work is finished, such as file processing requested during a software build.
We call the collection of requests handled in a shared computing environment over time a \emph{workload}.
In a task-by-task workflow, users launch separate agent sessions and remain responsible for coordinating their work.
We envision delegating this responsibility to a persistent agent system that manages an evolving workload on the user's computer.

The agent must construct each task's execution plan while deciding when to accept more work, how to delegate it, and which activities to overlap.
Task solving and workload management are therefore coupled.
Overlapping execution can use resources left idle during model calls or tool waits, but can also introduce contention and coordination overhead under shared CPU and memory constraints.
Performance on isolated tasks does not establish how well an agent maintains progress across requests, or whether its execution decisions improve workload delivery and efficiency.

\begin{figure}[t]
    \centering
    \includegraphics[width=\aosfigonewidth]{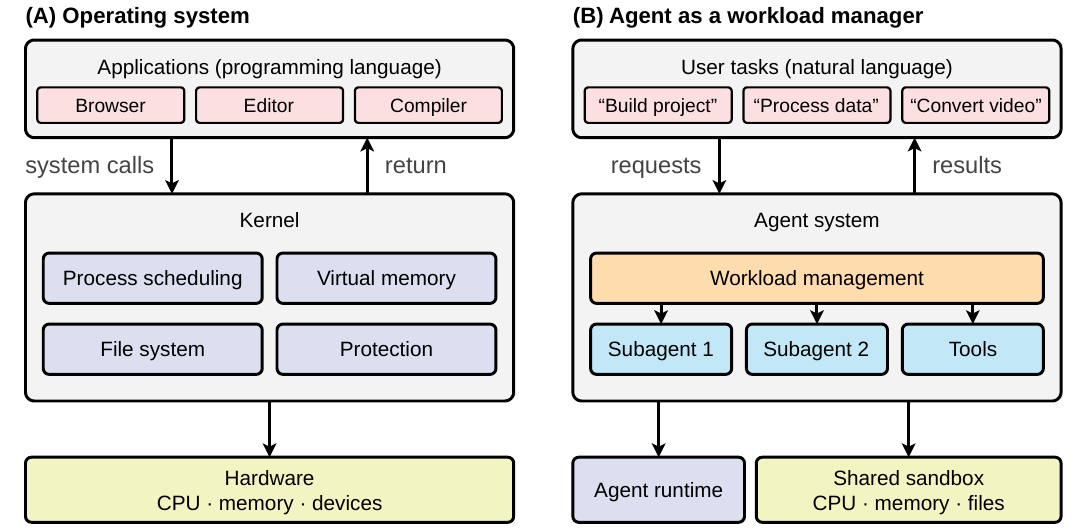}
    \caption{From programs to user tasks.
    (A) Applications written in programming languages issue system calls to an operating-system kernel, which provides process scheduling, virtual memory, file-system, and protection services over shared hardware.
    (B) User tasks expressed in natural language are submitted as requests to an agent system.
    Its workload management delegates work to subagents and tools, and the system runs on an agent runtime and a shared sandbox with CPU, memory, and files.
    ``Natural-language operating system'' is a metaphor for this job-management role.
    \benchname{} evaluates workload management and task execution.}
    \label{fig:workload-manager-comparison}
\end{figure}

This motivates our central question: \emph{Can current frontier agent systems effectively manage dynamic workloads under shared resource constraints?}
We call this role \emph{Agent as a Workload~Manager}.
Figure~\ref{fig:workload-manager-comparison} illustrates the role through a limited analogy with an operating system organizing applications.

We introduce \benchname{}\footnote{\csname benchname_explain\endcsname}, which makes a complete workload the unit of evaluation.
A persistent agent system handles tasks from existing CLI benchmarks in a shared environment under fixed CPU and memory limits.
To distinguish workload management from task execution, we evaluate each system in two settings: \emph{individual} (each task runs alone, providing an isolated capability baseline) and \emph{stream} (the same tasks arrive over time in a shared environment).
We compare verified delivery and execution efficiency across the two settings.

Our evaluation of five model--harness pairs shows that how much isolated task performance carries over to a shared workload differs widely across systems: every system loses 13.0 to 22.7 points on the heavy tier, and on the medium tier systems span only 11 points in individual SR but 26 points in stream SR, with no system holding the same rank in both orderings.
The sharpest case is a pair of systems whose individual baselines differ by 0.6 points on medium workloads (74.9\% and 74.3\%) yet deliver 53.9\% and 68.1\% on the same task set.
Execution-policy interventions identify task intake as a primary factor limiting delivery: serial dispatch leaves 73\% of heavy-tier tasks unclaimed while native management claims three quarters of them, and trial-level attribution finds task-solving failures in at most one scored trial per system, so most of the delivery gap is carried by orchestration outcomes.
Managing a workload well, however, is not the same as managing it cleverly: a fixed concurrency of two --- a constant with no workload reasoning --- comes within four points of the best agent-managed policy on both tiers, guidance raises the share of tasks claimed on heavy workloads while lowering delivery, and no policy exceeds 30\% median utilization of the four provisioned cores while tasks wait.
A thirty-task case study shows a single system spanning the full outcome range across identical repeats, suggesting that single-task-at-a-time execution can be a stochastic failure mode.

Our contributions are:
\begin{itemize}\aoscontribspacing
    \item \textbf{A formulation of agent workload management} and the \benchname{} benchmark with construction pipeline, execution protocol, and metrics.
    \item \textbf{A comparative evaluation} of five model--harness pairs showing that the share of isolated task performance a system retains under a shared workload differs widely across systems and reorders them on medium workloads.
    \item \textbf{Execution-policy interventions} identifying task intake as a primary driver of delivery gaps, and showing that avoiding serial execution captures most of the available gain while dynamic management yields no measurable advantage over a fixed concurrency constant.
\end{itemize}

\section{Evaluation Framework}
\label{sec:benchmark}

\begin{figure}[t]
    \centering
    \includegraphics[width=\linewidth]{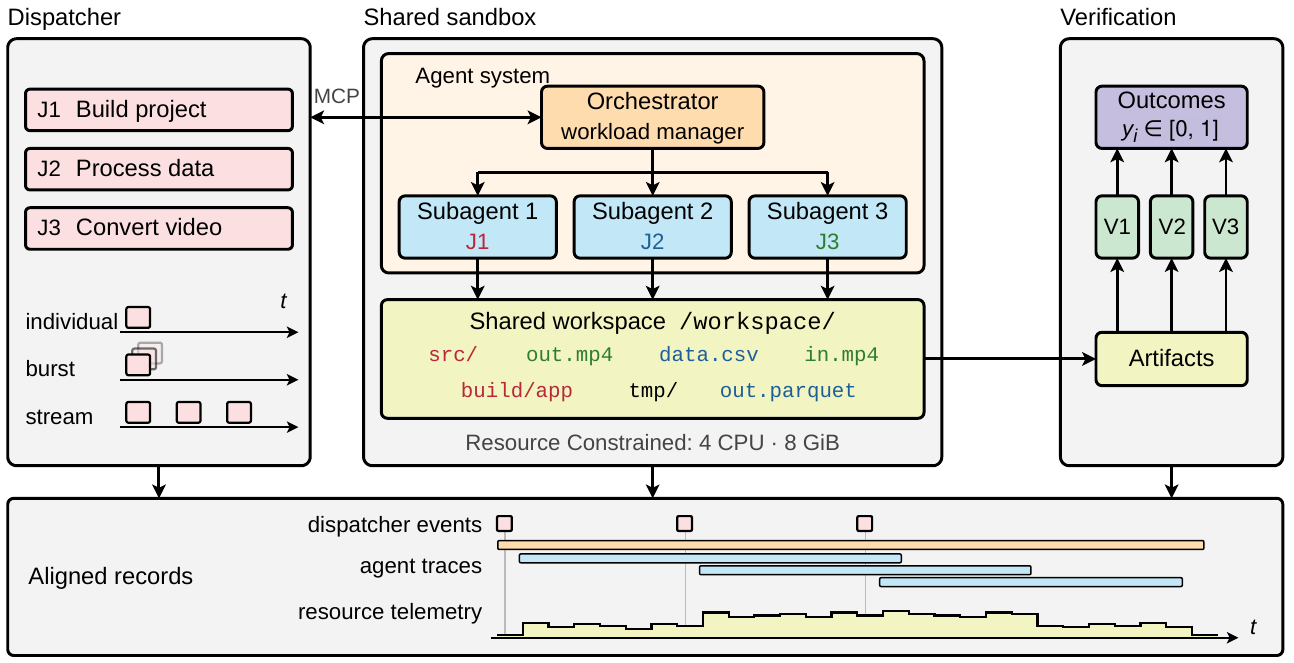}
    \caption{Evaluation framework.
    An external dispatcher releases tasks at once (\emph{burst}) or over time (\emph{stream}); the \emph{individual} baseline executes each task separately.
    Through MCP, the agent system can inspect all released tasks, claim work, and delegate execution within a shared 4-CPU, 8~GiB workspace.
    After execution ends, independent verifiers score final artifacts; dispatcher events, agent traces, and telemetry share a common clock.
    The illustrated delegation topology is not prescribed.}
    \label{fig:workload-evaluation}
\end{figure}

\benchname{} evaluates how well an agent system completes a \emph{workload}---a set of tasks released over time under a shared CPU and memory budget.
An \emph{agent system} is a model paired with an agent harness that provides its interaction loop, tools, and execution interface.
We do not prescribe internal organization: the system may execute requests in its main session, launch subagents, or invoke other tools.

\subsection{Task formulation}
\label{sec:formulation}

A workload is $\mathcal{W}=(\mathcal{J},B)$ where $\mathcal{J}=\{J_i\}_{i=1}^n$ and each task $J_i=(x_i,a_i,V_i)$ consists of a natural-language instruction $x_i$, a release time $a_i$, and a verifier $V_i$ that checks success against fixed criteria.
$B$ specifies the shared CPU and memory budget; one end-to-end run is a \emph{trial}.

Our primary setting is \emph{stream execution}, in which tasks arrive while the agent is already working.
An external dispatcher releases each task $J_i$ at its release time $a_i$; the agent system cannot see an instruction before its release, nor the schedule of future releases.
Through a Model Context Protocol (MCP) interface, the agent system can query the dispatcher to view all tasks released so far and their current states.
Using this view, the agent system decides when to take on additional work, how to organize and delegate execution, and when to check progress.
A task is \emph{claimed} once the agent system marks it as taken in the dispatcher; a claim only declares ownership and does not indicate that execution has started or succeeded.

We use two comparison settings in addition to stream execution.
\label{sec:conditions}
\emph{Individual} runs execute each task separately and provide a reference for isolated task-solving capability.
Because they collectively receive more resource opportunity than a shared run, they serve as a capability reference rather than an equal-cost throughput baseline.
\emph{Burst} runs use the same shared environment and task set as stream runs, but release all tasks at $t=0$.
In stream and burst runs, the agent system may still delegate each task to its own subagent, as in individual execution; the difference is that it must create and coordinate these subagents itself within the shared budget.
Figure~\ref{fig:workload-evaluation} illustrates the framework and its execution settings.

The protocol requires agent and worker execution to stop before verification.
The evaluator then runs every task verifier $V_i$ against the final state of the shared workspace.
\label{sec:verification}
Verification feedback is not exposed during execution.
Agent-reported \texttt{completed} and \texttt{failed} states are retained as behavioral records but do not determine correctness.
The default protocol scores all declared tasks, including tasks that were never claimed or never reported \texttt{completed}; tasks whose grader cannot judge an agent are excluded as described in Section~\ref{sec:setup}.
We additionally record task events, agent and tool traces, and container resource telemetry to characterize how the workload was executed.

\subsection{Workload construction}
\label{sec:construction}

\begin{figure}[t]
    \centering
    \includegraphics[width=\linewidth]{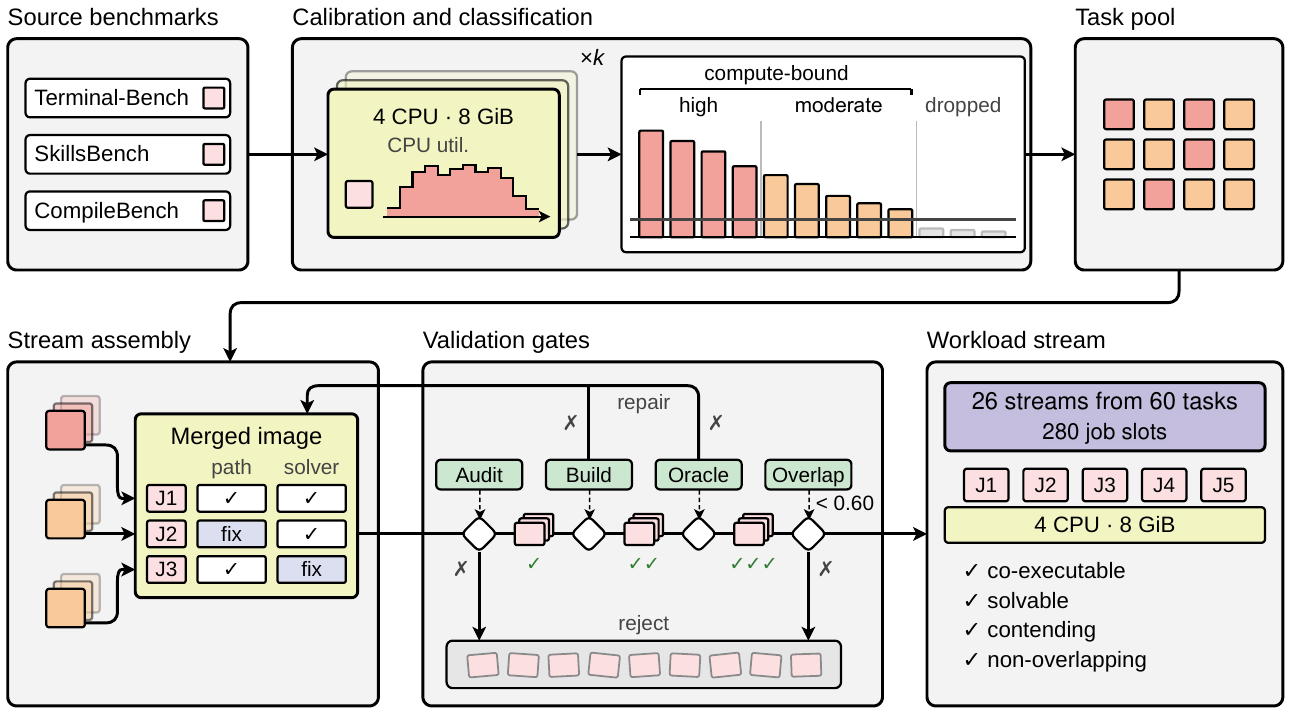}
    \caption{Workload construction pipeline.
    Source tasks are profiled for resource intensity (L2 moderate, L3 high), assembled into a shared Docker image with consistent path resolution, and admitted only after passing the validation gates.
    Full selection rules are in Appendix~\ref{app:data-construction}.}
    \label{fig:workload-synthesis}
\end{figure}

Figure~\ref{fig:workload-synthesis} summarises the construction pipeline.
Source benchmarks include Terminal-Bench~\citep{terminalbench}, SkillsBench~\citep{skillsbench}, TerminalWorld~\citep{terminalworld}, WildClawBench~\citep{wildclawbench}, CompileBench~\citep{compilebench}, CompileAgent~\citep{compileagent}, ParEval~\citep{pareval}, and the One Billion Row Challenge~\citep{onebrc}, supplemented by synthetic tasks.
Tasks are standardised in the Harbor format~\citep{harbor_task_format} with instructions, environments, reference solutions, and verifiers.
Calibration runs measure execution duration and CPU utilization; L2 and L3 denote moderate and high resource intensity (Appendix~\ref{app:data-construction}).

Task sets are merged into a shared Docker image with consistent path resolution across instructions, reference solutions, and verifiers.
Iterative validation applies four gates: a source audit, a \emph{Docker validity} gate (image builds and launches), a \emph{solution validity} gate (each reference solution must reach full score in the merged environment), and an overlap check against admitted workloads.
Failed candidates undergo repair and rechecking; admission requires passing all checks.

\subsection{Evaluation metrics}
\label{sec:metrics}

\begin{table}[t]
\centering
\caption{Workload-tier definitions. Outer limit is a harness termination, not a task deadline.}
\label{tab:workload-tiers}
\scriptsize\setlength{\tabcolsep}{3pt}
\begin{tabular}{lrrrrrr}
\toprule
Tier & Tasks & L2 & L3 & $n$ & Limit (min) \\
\midrule
Medium & 10 & 10 & 0 & 10 & 150 \\
Heavy  & 20 & 10 & 10 & 5  & 240 \\
\bottomrule
\end{tabular}
\end{table}

We report two main metrics, task success rate (SR) and the rewarded-task throughput ratio, which measure how well an agent system delivers verified work and how fast it does so relative to individual execution.
We also report total execution time (ET) and the memory-allocation overrun rate (labeled OOM in tables) as complementary resource metrics; lower ET reflects an efficiency gain only at comparable SR, since an agent system also finishes faster by leaving tasks unfinished.

\paragraph{Task success rate (SR).}
SR is the average reward across the scored tasks of a workload:
\begin{equation}
    \mathrm{SR}(\mathcal{W})=\frac{1}{|\mathcal{I}|}\sum_{i\in\mathcal{I}} y_i,
\end{equation}
where $\mathcal{I}$ is the set of scored tasks, $|\mathcal{I}|$ is its size, and $y_i\in[0,1]$ is the reward from task $i$'s verifier, with $y_i=1$ for full success, $y_i=0$ for no credit, and intermediate values for partial credit.
By default, $\mathcal{I}$ contains all $n$ tasks, including those never claimed or never finished; tasks whose verifier fails to execute are excluded rather than scored as zero (Section~\ref{sec:setup}).
Tables show $100\,\mathrm{SR}$ and $\Delta\mathrm{SR}=100(\mathrm{SR}_\mathrm{str}-\mathrm{SR}_\mathrm{ind})$; $\Delta\mathrm{SR}$ is an absolute difference between percentage-valued scores.
Eleven tasks are excluded from both settings (Appendix~\ref{app:exclusions}); in a stream trial whose sampled memory exceeds the 8~GiB allocation, tasks completed after the first such sample score zero.

\paragraph{Rewarded-task throughput ratio.}
Execution time (ET) covers model calls, tool use, coordination, and waiting for arrivals, excluding setup and post-run verification.
Stream ET is the shared run's elapsed time; the individual reference sums per-task execution times, including failed attempts, for sequential execution.
Let $Q_s=\sum_w\bar K_{s,w}/\sum_w\bar T_{s,w}$ for setting $s\in\{\mathrm{ind},\mathrm{str}\}$, where $\bar K_{s,w}$ and $\bar T_{s,w}$ are workload $w$'s total reward and ET averaged over repeats.
The throughput ratio is
\begin{equation}
  R_\mathrm{tput}=\frac{Q_\mathrm{str}}{Q_\mathrm{ind}}
  =\frac{\mathrm{SR}_\mathrm{str}}{\mathrm{SR}_\mathrm{ind}}
  \cdot\frac{\overline{\mathrm{ET}}_\mathrm{ind}}{\overline{\mathrm{ET}}_\mathrm{str}},
\end{equation}
where SR is matched over the same scored task mixture and $\overline{\mathrm{ET}}_s$ is mean ET per workload.
The ratio combines relative reward with relative execution time: 80\% reward in 100 minutes versus 60\% in 40 minutes gives $0.75\times2.5\approx1.9$.
A value above one indicates more rewarded work per minute than the system's own sequential individual reference.
We aggregate rewards and times before dividing, rather than averaging workload ratios (Appendix~\ref{app:metric-derivations}).

\paragraph{Total execution time (ET).}
ET covers the agent execution phase: model calls, tool use, coordination, and waiting for arrivals, but excludes environment setup and post-run verification.
Stream ET is the shared run's elapsed wall-clock time; individual ET is the sum of per-task execution times (including failed attempts) that constitute the sequential reference.
We report mean ET per workload; execution boundaries for the individual reference are estimated from per-task timestamps.

\paragraph{OOM rate.}
The OOM rate counts the share of stream trials in which any memory sample exceeds the 8~GiB allocation; tasks completed after the first such sample receive no credit in SR.
The allocation is not kernel-enforced in these containers, so the rate measures excess demand rather than observed terminations (Appendix~\ref{app:setup-details}).

\paragraph{Matched aggregation.}
For each stream trial, the individual reference is the mean individual reward over that trial's own scored tasks, so both sides score the same task mixture.
Metrics are first averaged within workloads, then equally across workloads; repeating a workload does not increase its weight.

\section{Experimental Results}
\label{sec:results}

\subsection{Setup}
\label{sec:setup}

\paragraph{Systems and tiers.}
We evaluate five model--harness pairs: Claude Opus 5~\citep{claude_opus_5}, GLM-5.3~\citep{glm5}, and DeepSeek-V4-Pro~\citep{deepseek_v4_pro} with Claude Code~\citep{claude_code}; GPT-5.6-sol~\citep{gpt56sol} with Codex~\citep{openai_codex}; and Kimi-K3~\citep{kimi_k3} with Kimi Code~\citep{kimi_code}.
Both Claude Code and Codex use the \texttt{high} reasoning-effort setting (versions in Appendix~\ref{app:setup-systems}).
We report medium (10 tasks, 10 workloads) and heavy (20 tasks, 5 workloads) tiers; a thirty-task workload is used only as a case study (Section~\ref{sec:hplus-case}).

\paragraph{Execution and scoring.}
Each run is allocated 4 CPUs and 8~GiB, shared in stream and exclusive in individual execution.
We target three attempts per workload and per task and retain the first three eligible repeats under the \emph{full-clean} filter: reward record, transcript, ${\geq}8$ trajectory steps, and no output-cap, throttling, dead-route, or verifier-fetch failure.
Individual SR is matched to each stream trial's scored tasks and aggregated like stream SR, within workloads and then equally across workloads; task matching uses source identifier and instruction SHA-256 hash.
Eleven tasks are excluded from both settings (Appendix~\ref{app:exclusions}): ten that produced at least one false zero, a reward of $0$ written by a grader that never ran, and one whose LLM-judge grader cannot run in the sandbox; retained scoring sets contain 87 and 79 of the 100 task slots in medium and heavy.
GPT-5.6-sol medium stream is incomplete at 27 of 30 trials due to an image-level setup failure (Appendix~\ref{app:codex-setup}).
CPU estimates come from cumulative cgroup counters aligned to execution-phase timestamps; in a stream trial whose sampled memory exceeds the 8~GiB allocation, tasks completed after the first such sample score $0$.
SR differences are paired and use a cluster bootstrap over workloads (4000 resamples); medium has ten clusters and heavy five.

\subsection{Main Results}
\label{sec:main-results}

Table~\ref{tab:main-results} reports task-matched individual and stream SR alongside the rewarded-task throughput ratio and total execution time (ET) for the medium and heavy tiers.
Individual ET is the sum of per-task execution times averaged across workloads; stream ET is the elapsed shared execution interval.

\begin{table}[t]
\caption{Individual and stream results.
SR is mean per-task reward; $\Delta$ = Str.\ $-$ Ind.\ (\%); individual SR is taken over each stream trial's scored tasks, matched by task id and instruction hash; both are averaged within workloads, then equally across workloads.
Eleven tasks are excluded from both sides: ten that produced at least one \emph{false zero} --- a reward of $0$ written by a grader that never ran, identified by the absence of any test-result line in its output; one of them also fails its own reference solution --- and one whose LLM-judge grader cannot run (Appendix~\ref{app:exclusions}).
In a stream trial whose sampled memory exceeds the 8~GiB allocation, tasks completed after the first such sample score $0$; OOM is the share of that system's stream trials with such an overrun.
ET is agent execution time; Ind.\ is the sequential cost of the same task set, so Throughput $= (\mathrm{SR}_{\mathrm{str}}/\mathrm{SR}_{\mathrm{ind}})\times(\mathrm{ET}_{\mathrm{ind}}/\mathrm{ET}_{\mathrm{str}})$ is rewarded work per execution minute relative to sequential.
Rate is the absolute stream delivery rate, rewarded tasks per execution hour.
Str.\ ET also counts time spent on the excluded tasks; all other columns use the same scored task set. Bold = best per tier for $\Delta$, Throughput and Rate.}
\label{tab:main-results}
\centering
\footnotesize
\setlength{\tabcolsep}{3pt}
\renewcommand{\arraystretch}{1.04}
\begin{tabular*}{\linewidth}{@{\extracolsep{\fill}}llrrrrrrrr@{}}
\toprule
Model & Harness & \multicolumn{3}{c}{SR (\%)} & Thrpt. & \multicolumn{2}{c}{ET (min)} & Rate & OOM \\
\cmidrule(lr){3-5}\cmidrule(lr){7-8}
& & Ind. & Str. & $\Delta$ & ($\times$) & Ind. & Str. & (/h) & (\%) \\
\midrule
\multicolumn{10}{@{}l}{\textbf{Medium} (10 workloads, 20 minutes of arrivals, 150-minute limit)} \\
Claude Opus 5 & Claude Code & 81.5 & 79.7 & \textbf{$-$1.8} & 2.06 & 213.2 & 101.7 & 4.09 & 3.4 \\
GLM-5.3 & Claude Code & 74.9 & 53.9 & $-$21.1 & 1.73 & 359.7 & 148.8 & 1.88 & 0.0 \\
DeepSeek-V4-Pro & Claude Code & 79.5 & 54.4 & $-$25.1 & 1.49 & 321.2 & 147.4 & 1.92 & 0.0 \\
Kimi-K3 & Kimi Code & 74.3 & 68.1 & $-$6.3 & \textbf{2.22} & 295.2 & 122.4 & 2.91 & 3.3 \\
GPT-5.6-sol & Codex & 85.3 & 70.9 & $-$14.5 & 1.68 & 129.2 & 63.9 & \textbf{5.74} & 0.0 \\
\midrule
\multicolumn{10}{@{}l}{\textbf{Heavy} (5 workloads, 60 minutes of arrivals, 240-minute limit)} \\
Claude Opus 5 & Claude Code & 75.5 & 62.5 & \textbf{$-$13.0} & 1.85 & 400.7 & 179.9 & \textbf{3.29} & 20.0 \\
GLM-5.3 & Claude Code & 60.1 & 44.4 & $-$15.7 & \textbf{2.46} & 722.4 & 217.8 & 1.95 & 13.3 \\
DeepSeek-V4-Pro & Claude Code & 58.1 & 35.4 & $-$22.7 & 1.57 & 599.8 & 240.1 & 1.44 & 6.7 \\
Kimi-K3 & Kimi Code & 65.3 & 46.0 & $-$19.3 & 1.77 & 541.2 & 213.6 & 2.02 & 40.0 \\
GPT-5.6-sol & Codex & 71.0 & 52.9 & $-$18.1 & 1.04 & 245.6 & 174.2 & 2.81 & 0.0 \\
\bottomrule
\end{tabular*}
\end{table}

\begin{figure}[t]
\centering
\begin{minipage}[t]{0.49\linewidth}
\centering
\includegraphics[width=\linewidth]{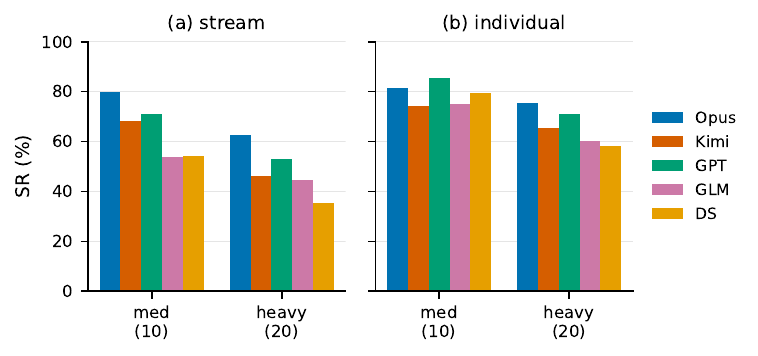}
\caption{Mean per-task reward for the five systems under individual and stream settings.
Labels: opus = Claude Opus 5, gpt = GPT-5.6-sol, ds = DeepSeek-V4-Pro; tiers are medium (10 tasks) and heavy (20 tasks), with partially overlapping task sets.
Comparisons are within-tier only; the individual setting uses no scheduling preamble (Appendix~\ref{app:setup-prompt}).}
\label{fig:main-results}
\end{minipage}\hfill
\begin{minipage}[t]{0.49\linewidth}
\centering
\includegraphics[width=\linewidth]{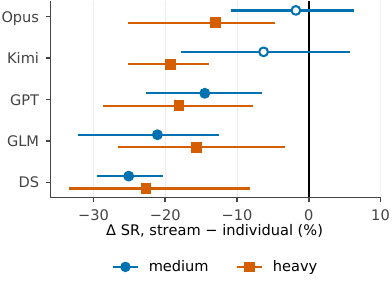}
\caption{Stream-minus-individual difference $\Delta$ of Table~\ref{tab:main-results} with 95\% workload-clustered bootstrap intervals (4000 resamples).
Filled markers exclude zero; heavy has five clusters, so its intervals are a coarse check rather than calibrated.}
\label{fig:paired-diff}
\end{minipage}
\end{figure}

\paragraph{On medium workloads, how much ability is retained separates the systems more than ability itself.}
On heavy workloads Claude Opus 5 delivers 62.5\% against a 75.5\% individual baseline ($\Delta = -13.0$), whereas the other four lose 15.7 to 22.7 points (retaining 61\%--75\% of their baselines); on medium Opus loses 1.8 points (98\% retention) and the others 6.3 to 25.1 (68\%--92\%).
On heavy every loss is significant (all paired intervals in Figure~\ref{fig:paired-diff} exclude zero); on medium the two smallest, Opus's and Kimi-K3's, are not distinguishable from zero. What separates the systems is how much they lose.
The spread is not a spread in task-solving ability.
GPT-5.6-sol has the highest medium baseline (85.3\%) yet gives up 14.5 points, while Kimi-K3, whose medium baseline is 11.0 points lower, gives up only 6.3.
DeepSeek-V4-Pro shows the same pattern more sharply: a medium baseline of 79.5\%, third of the five, yet the tier's second-lowest stream SR at 54.4\%.
Individual ability therefore does not determine workload delivery on medium (Spearman $\rho=0.6$ between the two orderings, $n=5$): the medium baselines span 11.0 points across the five systems while their stream scores span 25.8 points, and no system holds the same rank in both orderings on that tier; on heavy the two orderings coincide.
The quantity that separates the systems is what they do with the workload, not what they can do with a task.
Appendix~\ref{app:intake-collapse} traces part of the heavy-tier gap to a specific behaviour: GLM-5.3 in some runs and DeepSeek-V4-Pro in every run stop claiming released tasks and execute one at a time, leaving 25\% and 48\% of released tasks never claimed against 0\% for Claude Opus 5.

\paragraph{Most of the heavy-tier gap is lost outside task execution.}
Splitting each heavy-tier loss by how the task ended, tasks never claimed, still unfinished at the limit, or completed only after a memory overrun account for 7.3 of Claude Opus 5's 13.0 points and for 16.4 to 25.6 points of the other four systems, whose losses are 15.7 to 22.7 points, while tasks the system completed normally contribute at most 1.3 points net (Appendix~\ref{app:intake-collapse}).
The same holds on medium for GLM-5.3 and DeepSeek-V4-Pro, whereas for GPT-5.6-sol about half of the medium loss (6.8 of 14.5 points) comes from tasks it completed.

\paragraph{Bootstrap intervals resolve the larger decreases.}
Workload-clustered 95\% intervals (4000 resamples, Figure~\ref{fig:paired-diff}) on medium exclude zero for DeepSeek, GLM, and GPT-5.6-sol at $[-29.4, -20.5]$, $[-32.0, -12.7]$, and $[-22.6, -6.7]$ points; Kimi and Opus do not ($[-17.7, 5.6]$ and $[-10.7, 6.1]$).
Heavy has five clusters and is a coarse check only; there all five intervals exclude zero.

\paragraph{Repeated instances can bias stream scores upward.}
Workloads contain multiple instances of the same task sharing one filesystem, so an artifact one instance produces can satisfy a sibling's verifier; isolated execution cannot do this.
This biases stream SR upward wherever families occur, so the heavy decreases are conservative.
Only heavy workloads repeat a task, and there the 42 cells whose task has a sibling instance in the same workload show stream-minus-individual differences of $+7.1$ to $-16.2$ points, against $-15.3$ to $-25.5$ for singleton tasks, in every system.
The repeated tasks also differ in kind from the singletons, so this does not isolate artifact reuse, but it is the direction reuse predicts.
Tasks on which a system's individual baseline is zero are kept in Table~\ref{tab:main-results}; dropping them makes every $\Delta$ 2.3 to 11.1 points more negative and moves GLM-5.3 from second to fourth on heavy (Appendix~\ref{app:zero-baseline}).

\paragraph{Throughput distinguishes delivery from shorter execution.}
On medium workloads the rewarded-task throughput ratios are $2.22\times$ for Kimi, $2.06\times$ for Opus, $1.73\times$ for GLM, $1.68\times$ for GPT-5.6-sol, and $1.49\times$ for DeepSeek.
All five exceed unity, so every system delivers rewarded work faster than sequential individual execution even where its per-task reward drops.
The ordering is not the ordering of $\Delta$: Opus loses the least reward per task yet Kimi has the higher ratio, because Opus's own sequential reference is already the faster of the two.
GPT-5.6-sol's lower ratio comes from its SR loss instead: its time ratio ($2.02\times$) matches Opus's ($2.10\times$), but it keeps only 83\% of its baseline SR.
The ratio therefore measures delivery rate against a system's own sequential reference, not cross-system speed.
For cross-system speed, Table~\ref{tab:main-results} also gives the absolute rate: GPT-5.6-sol delivers the most rewarded tasks per hour on medium (5.74) and Claude Opus 5 on heavy (3.29).
The absolute rate also reflects how fast each model endpoint responds, so it does not isolate workload management.
On heavy the ratios fall for three of five systems and rise only for GLM ($1.73\times \to 2.46\times$) and DeepSeek ($1.49\times \to 1.57\times$); GPT-5.6-sol drops to $1.04\times$, barely above sequential.

\paragraph{Execution time and memory demand.}
Stream execution reduces total execution time (ET) per workload by factors of 1.4--3.3 in every cell; the reductions do not track SR losses (on medium, GLM and DeepSeek cut ET comparably to Opus while losing 19.3 and 23.3 more points), so these are not equivalent-output efficiency gains.
Memory overruns concentrate on heavy: 40.0\% of Kimi's stream trials and 20.0\% of Opus's exceed the 8~GiB allocation, against a single overrunning trial each for Claude Opus 5 and Kimi-K3 on medium and none for the other three systems; tasks completed after an overrun are not credited (Appendix~\ref{app:setup-details}).

\section{Analysis}
\label{sec:analysis}

\subsection{Execution organization and workload delivery}

We compare GLM-5.3 with Claude Code under four execution policies: (i) native agent management, (ii) native with additional resource and intake guidance, (iii) serial FIFO ($k=1$), and (iv) fixed concurrency of two tasks ($k=2$).
Fixed policies use the same model and solver but dispatch via a driver; task sets, arrival schedules, and resource limits are matched.

\begin{figure}[t]
\centering
\begin{minipage}[t]{0.54\linewidth}
\centering
\scriptsize
\setlength{\tabcolsep}{2.5pt}
\captionof{table}{GLM-5.3 execution-policy comparisons under stream arrivals.
$k$ limits unfinished solver sessions. Claim and Compl.\ are the shares of scored tasks claimed and reported complete; $t_{\text{claim}}$ is median seconds from claimable to claimed; ET and CPU are per-trial medians. All columns use the task set and scoring of Table~\ref{tab:main-results}, so the Native rows equal its GLM-5.3 stream SR.}
\label{tab:policy-results}
\begin{tabular}{llrrrrrrr}
\toprule
Tier & Policy & $N$ & SR & ET & Claim & Compl. & $t_{\text{claim}}$ & CPU \\
& & & (\%) & (min) & (\%) & (\%) & (s) & (min) \\
\midrule
Med & Serial ($k{=}1$) & 30/30 & 39.5 & 150.0 & 51.7 & 40.6 & 3360 & 98.5 \\
Med & Native & 30/30 & 53.9 & 150.1 & 75.9 & 60.9 & 1204 & 93.4 \\
Med & Guidance & 30/30 & 60.2 & 150.0 & 81.2 & 67.0 & 1642 & 134.0 \\
Med & Fixed ($k{=}2$) & 30/30 & 56.6 & 150.0 & 80.1 & 59.4 & 2614 & 181.1 \\
\midrule
Hvy & Serial ($k{=}1$) & 15/15 & 28.6 & 240.0 & 30.0 & 23.6 & 4819 & 189.4 \\
Hvy & Native & 15/15 & 44.4 & 240.1 & 75.9 & 59.5 & 1190 & 268.7 \\
Hvy & Guidance & 15/15 & 39.2 & 240.1 & 86.1 & 63.7 & 2502 & 318.2 \\
Hvy & Fixed ($k{=}2$) & 15/15 & 42.8 & 240.0 & 53.6 & 43.5 & 3600 & 294.1 \\
\bottomrule
\end{tabular}
\end{minipage}\hfill
\begin{minipage}[t]{0.44\linewidth}
\small
\paragraph{Serial intake caps delivery.}
Serial FIFO has the lowest SR (39.5\%/28.6\%) and completion share (40.6\%/23.6\%), with claim latency three to four times native; on heavy 73.3\% of all released tasks are never claimed.

\paragraph{A fixed concurrency constant comes close to agent management.}
Fixed $k=2$ scores 2.7 points above native on medium and 1.6 points below it on heavy, within four points of the best policy in both tiers, although it claims only half of the heavy tasks.

\paragraph{More intake need not mean more delivery.}
Guidance raises coverage on both tiers but SR only on medium; on heavy SR falls 5.2 points while the share of tasks claimed rises 10.2.
Serial heavy at a 10-hour limit recovers only 26\% of its gap to native despite 2.5$\times$ the budget, so additional time alone does not close the serial deficit.
\end{minipage}
\end{figure}

\begin{figure}[t]
\centering
\begin{minipage}[t]{0.49\linewidth}
\centering
\includegraphics[width=\linewidth]{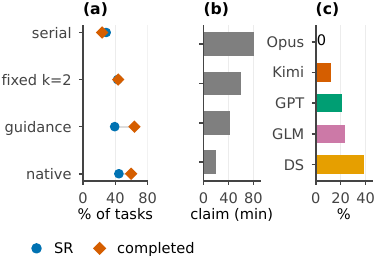}
\caption{(a) SR and share of tasks completed for the four GLM-5.3 policies on heavy workloads. (b) Their median claim latency. (c) Share of released tasks never claimed, per system, pooled over medium and heavy stream trials.}
\label{fig:mechanism}
\end{minipage}\hfill
\begin{minipage}[t]{0.49\linewidth}
\centering
\includegraphics[width=\linewidth]{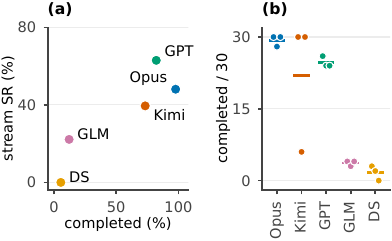}
\caption{Thirty-task case study, scored as in Table~\ref{tab:main-results}. (a) Stream SR vs.\ share of released tasks completed. (b) Per-trial completion counts (three repeats per system; bar = mean). Kimi-K3 spans 30, 6, and 30 completions across identical repeats.}
\label{fig:hplus}
\end{minipage}
\end{figure}

\subsection{Arrival conditions}

The GLM heavy burst comparison attains 29.9\% SR against 44.4\% for stream arrivals: releasing every task at once pushes memory past the allocation in 7 of 15 trials, against 2 of 15 under stream arrivals.
Claim latency separates sharply (327~s burst vs.\ 1190~s stream), so faster intake does not translate into delivery when the work taken on at once exceeds the memory allocation and later completions go uncredited.

\subsection{A thirty-task workload as a case study}
\label{sec:hplus-case}

A thirty-task workload, three repeats per system, stresses intake harder than either reported tier and exposes the mechanism in isolation.
Stream SR spans 0.0\% to 63.0\% against individual baselines of 55.6\% to 72.2\%.
Opus claims all 90 task instances across its three repeats and completes 88; GPT-5.6-sol completes 74 while never holding more than two tasks at a time, so fast serial execution can still cover most of the workload.
GLM and DeepSeek complete 11 and 5, both holding never more than one task at a time and reaching the 360-minute limit with 76 and 82 instances never claimed.
Both systems score 63.6\% on the individual baseline, so the deficit is not inability to solve the tasks.

\paragraph{The failure mode can be stochastic.}
Kimi-K3 completes 30, 6, and 30 tasks across identical repeats --- the second repeat held one task at a time, reproducing the GLM/DeepSeek pattern, while the other two claimed the full workload.
A single system therefore spans both regimes on identical inputs, which suggests that serial-execution behavior can be a failure mode rather than a stable capability level.
Opus is the only system that claims the full workload in all three repeats.

Trajectory evidence and terminal-state decompositions are in Appendix~\ref{app:trajectory-detail} and~\ref{app:terminal-states-detail}.

\begin{figure}[t]
\centering
\begin{minipage}[t]{0.49\linewidth}
\centering
\includegraphics[width=\linewidth]{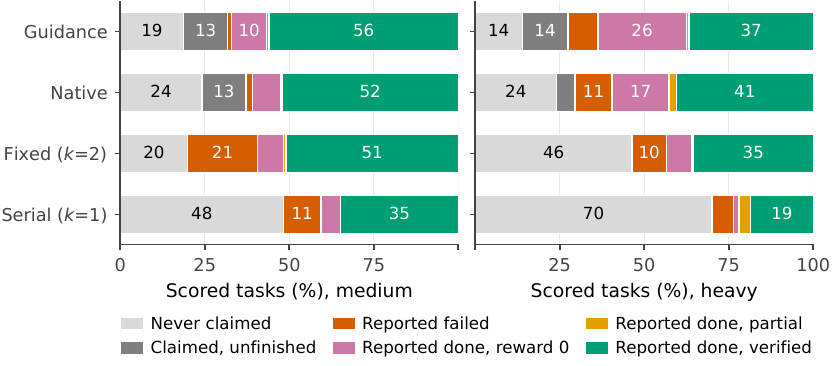}
\caption{Terminal states of the scored tasks of Table~\ref{tab:main-results} for the four GLM-5.3 policies.
Greys: never served; warm: served without full credit (partially solved tasks earn partial SR credit); green: verified.}
\label{fig:outcome-decomposition}
\end{minipage}\hfill
\begin{minipage}[t]{0.49\linewidth}
\centering
\includegraphics[width=\linewidth]{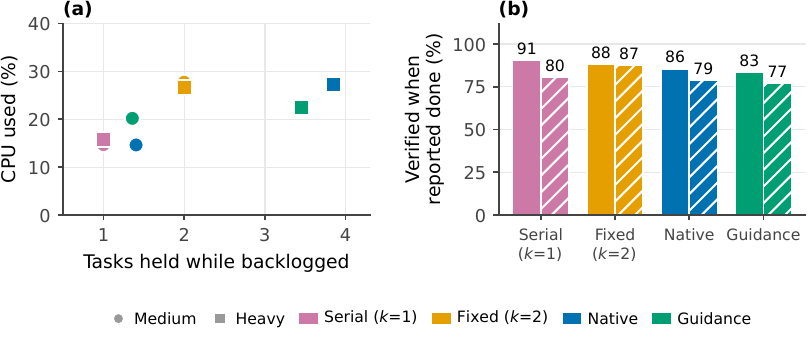}
\caption{(a) Tasks held vs.\ CPU utilization; (b) fraction of reported-complete tasks verified.
No policy exceeds 30\% median CPU while tasks wait; agent-managed policies have lowest verification accuracy.}
\label{fig:held-calibration}
\end{minipage}
\end{figure}

\begin{figure}[t]
\centering
\begin{minipage}[t]{0.49\linewidth}
\centering
\includegraphics[width=\linewidth]{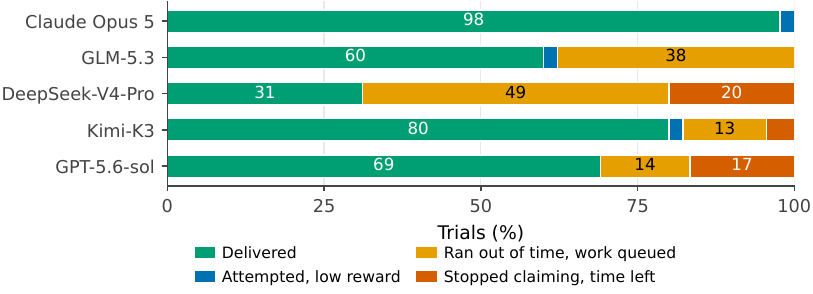}
\caption{Trial disposition of the 221 medium and heavy stream trials scored in Table~\ref{tab:main-results}.
Attempted-and-failed is rare ($\leq$1 per system); orchestration failures carry most of the delivery gap (Section~\ref{sec:main-results}).}
\label{fig:failure-buckets}
\end{minipage}\hfill
\begin{minipage}[t]{0.49\linewidth}
\centering
\includegraphics[width=\linewidth]{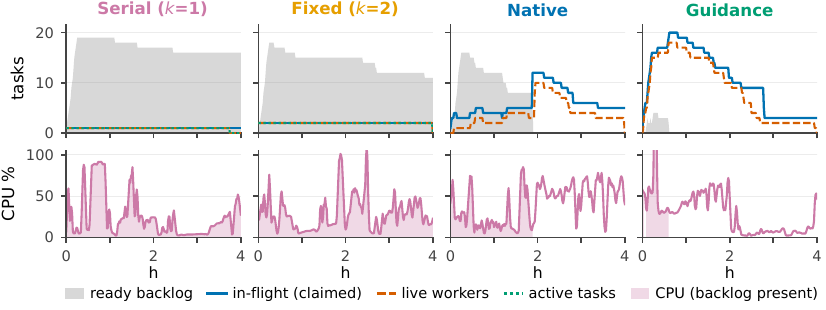}
\caption{One heavy workload under the four policies on a shared clock.
Serial and fixed $k{=}2$ leave 16 and 11 of 20 tasks unclaimed; agent-managed policies drain the backlog within two hours.}
\label{fig:timeline}
\end{minipage}
\end{figure}

\section{Limitations}
\label{sec:limitations}

The evaluation setting bounds what our results can mean. A task consists of an
instruction, a release time, and a verifier, with no deadline, no value or
priority, and no dependency on another task. The dispatcher releases work
first-in, first-out, so a system cannot choose which released task to serve
next. We therefore measure intake rate and execution organization, but never
the ordering policy that scheduling theory treats as the central decision, and
because every task is worth the same, the objective reduces to maximizing
total reward under a shared budget. The stream and individual settings also
differ in more than scheduling. The stream setting adds a shared filesystem, a
task-pool interface, a scheduling preamble, and contention for one allocation,
so the paired difference measures the cost of moving a task from isolation into
a managed workload rather than the cost of scheduling decisions alone.
Individual runs additionally receive more aggregate resource opportunity, which
makes them a capability reference rather than an equal-cost baseline. The
workload design samples one region of a larger space. Tiers of 10, 20, and 30
tasks over 4 CPUs and 8~GiB draw different but overlapping task sets (22 of medium's 29 scored task
variants recur in heavy and reuse the same individual baselines), so tier
differences confound scale with composition and the two tiers' estimates are
not independent, and one burst condition does not characterize
sensitivity to the arrival process. Whether these failure modes persist under
tighter resource ratios, or under hundreds rather than tens of concurrent
requests, is outside what this design establishes. Inference is finally bounded
by workloads rather than trials, since repeats share a task set and an
individual baseline. That leaves ten clusters on medium and five on heavy, too
few for a calibrated interval, and most heavy trials stop at the wall-clock
limit, so delivery on that tier is a lower bound.

\section{Conclusion}
\label{sec:conclusion}

We introduced \benchname{} to evaluate agent systems as workload managers,
composing verified CLI tasks into shared container workloads with controlled
arrivals and measuring five model--harness pairs in individual and stream
settings. The central finding is that how much isolated task performance
carries over to a shared workload differs widely across systems: every system
loses 13.0 to 22.7 points on heavy workloads, and on medium workloads the
systems span 11.0 points in
individual success rate but 25.8 in stream success rate with no system holding
its rank across both, and two systems whose individual baselines differ by 0.6
points deliver 53.9\% and 68.1\%. Execution-policy interventions
identify task intake as a primary factor limiting delivery --- serial execution cannot claim released work fast
enough to deliver a competitive share within the budget, while a fixed
concurrency of two comes within four points of the best agent-managed policy
on both tiers, so most of the available gain comes from not executing
serially --- and on the thirty-task
case a single system spans the full range of outcomes across identical
repeats, which suggests that single-task-at-a-time execution can be a
failure mode rather than a stable capability level. Workload management is
therefore a distinct evaluation dimension from task solving, and the immediate
work it calls for is enforced resource limits, instruction-level deduplication
for repeated task families, more heavy-tier workloads for calibrated
intervals, and trajectory-level attribution of execution overlap.

\section*{Ethics statement}
\benchname{} uses tasks from existing benchmarks and runs agent-written commands in containers shared by the tasks in each trial.
Release requirements include documenting upstream licenses, excluding secrets and private data, and describing the risks of executing untrusted code.

\section*{Reproducibility statement}
We will release the workload manifest, container and agent image identifiers, task and verifier provenance, prompts, and launcher configurations, together with the record of included trials and the scripts that compute every reported statistic and figure from it.
The scoring rules, exclusions and trial-selection criteria are stated in Section~\ref{sec:setup} and Appendix~\ref{app:exclusions}; resource measurements are estimates from sampled cgroup counters.

\section*{AI use statement}
The authors used generative AI tools for research planning, feedback on methods and experiments, software development and review, data inspection, and manuscript drafting and editing.
The authors reviewed the resulting materials and interpretations.
They take responsibility for the methods, results, claims, code, and text.

\setlength{\bibsep}{3pt}
{\Urlmuskip=0mu plus 1mu\relax
\bibliography{references}}
\bibliographystyle{iclr2027_conference}

\clearpage
\appendix
\section{Details of Data Constructions}
\label{app:data-construction}

\paragraph{Task sources.}
Task sources include Terminal-Bench \citep{terminalbench}, SkillsBench \citep{skillsbench}, TerminalWorld \citep{terminalworld}, WildClawBench \citep{wildclawbench}, CompileBench \citep{compilebench}, CompileAgent \citep{compileagent}, ParEval \citep{pareval}, and the One Billion Row Challenge \citep{onebrc}.
We also include supplementary synthetic tasks developed for \benchname{}.

\begin{figure}[htbp]
\centering
\includegraphics[width=\linewidth]{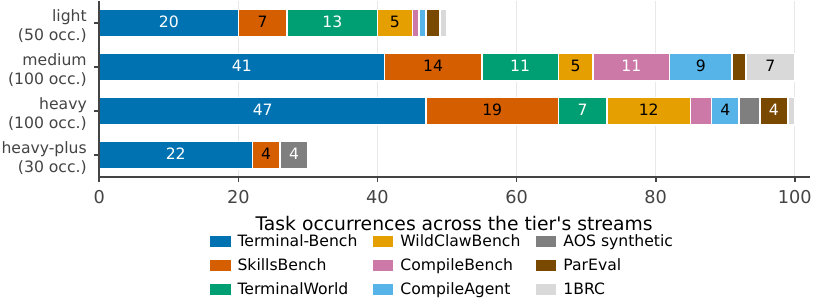}
\caption{Source-benchmark composition of the 26 evaluated stream workloads,
as task occurrences across each tier's streams (10, 10, 5, and 1 workloads).
Segments under four occurrences are unlabeled. Occurrences exceed distinct
tasks because a stream may release repeated instances of one task.}
\label{fig:stream-sources}
\end{figure}

Figure~\ref{fig:stream-sources} gives the resulting composition of the
evaluated workloads.
Terminal-Bench is the largest contributor in every tier, at 40\% to 73\% of
task occurrences, so the workloads are not balanced across sources and a
per-source effect cannot be separated from a per-tier one.
The counts are task \emph{occurrences} summed over each tier's streams, which
is the work an agent had to serve; the distinct-task counts are lower because
streams may release repeated instances of one task, at 21, 34, 48, and 19
distinct tasks for light, medium, heavy, and heavy-plus against 50, 100, 100,
and 30 occurrences.
Heavy-plus draws on three sources only, being a single workload.

\paragraph{Task eligibility and calibration.}
Source tasks must provide instructions, input assets, executable verifiers, and environments compatible with the shared container configuration.
The task pool uses two valid calibration runs per task with GLM-5.2~\citep{glm5} under 4 CPUs and 8 GiB of memory.
Calibration uses a different model from any evaluation system, so tier labels are not tuned to the systems later measured against them.
Valid calibration requires usable execution and resource records; it does not imply verification success.
Reference-solution validation separately checks solvability after merging.

\paragraph{Resource score and tier boundaries.}
For task $i$ and calibration run $k$, let $w_{ik}$ be elapsed time in seconds and $u_{ik}$ be mean CPU utilization divided by the allocated CPU capacity, so that $u_{ik}=1$ denotes full use of the quota.
We compute the resource score as $s_i=\frac{1}{2}\sum_{k=1}^{2}\min(w_{ik}/3600,1)\min(u_{ik},1)$.
The classification assigns L1 to $s_i<0.009$, L2 to $0.009\leq s_i<0.10$, and L3 to $s_i\geq0.10$.
L1, L2, and L3 denote low, moderate, and high observed resource intensity, respectively.
Workload construction uses tasks from L2 and L3.
Tier labels are those the construction pipeline assigned from these thresholds.

\paragraph{Workload overlap.}
For the task sets $\mathcal{J}_a$ and $\mathcal{J}_b$ of two workloads, we define overlap as $|\mathcal{J}_a\cap\mathcal{J}_b|/\min(|\mathcal{J}_a|,|\mathcal{J}_b|)$ and require it to be below $0.60$ against each admitted workload.
The smaller-set denominator limits near-containment of smaller workloads in larger ones.
This criterion limits task reuse without guaranteeing statistical independence.

\subsection{Stream data format}
\label{app:stream-format}

Our stream format is adapted from the Harbor task format \citep{harbor_task_format}.
A Harbor task packages an instruction in \texttt{instruction.md}, configuration in \texttt{task.toml}, an execution environment in \texttt{environment/}, a reference solution in \texttt{solution/}, and a verifier in \texttt{tests/}.
We extend this organization by placing task-specific instructions and verifiers under \texttt{tasks/<task\_id>/}, merging the environments, and replacing the top-level task configuration with \texttt{stream.toml}.
Figure~\ref{fig:stream-format} shows a minimal two-task example; its names and release times are illustrative.

\begin{figure}[htbp]
\centering
\begin{minipage}[t]{0.47\linewidth}
\textbf{Directory layout}
\small
\begin{verbatim}
example-stream/
|-- stream.toml
|-- environment/
|   |-- Dockerfile
|   `-- _sources/...
|-- tasks/
|   |-- build-project/
|   |   |-- instruction.md
|   |   `-- tests/
|   |       |-- test.sh
|   |       `-- ...
|   `-- process-data/
|       |-- instruction.md
|       `-- tests/
|           |-- test.sh
|           `-- ...
`-- solution/
    |-- solve.sh
    `-- tasks/<task_id>/...
\end{verbatim}
\end{minipage}\hfill
\begin{minipage}[t]{0.50\linewidth}
\textbf{\texttt{stream.toml}}
\small
\begin{verbatim}
version = "1.0"

[metadata]
name = "example-stream"

[environment]
cpus = 4
memory_mb = 8192
allow_internet = true

[session]
max_steps = 360

[[arrivals]]
task_id = "build-project"
arrive_wall_time = 0

[[arrivals]]
task_id = "process-data"
arrive_wall_time = 60
\end{verbatim}
\end{minipage}
\caption{Stream packaging adapted from Harbor tasks.
One environment serves both tasks; separate task directories retain their instructions and verification assets.
The configuration adds timed task releases to the shared environment specification.}
\label{fig:stream-format}
\end{figure}

\paragraph{Configuration fields.}
\texttt{version} identifies the format version, and \texttt{metadata.name} identifies the workload.
The \texttt{environment} block retains Harbor's environment configuration fields; its CPU and memory limits apply to the shared container, and \texttt{docker\_image} may additionally specify a prebuilt image.
The \texttt{session.max\_steps} field stores session-level execution metadata; its enforcement depends on the configured harness.
Each \texttt{arrivals} entry links a \texttt{task\_id} to its task directory and an \texttt{arrive\_wall\_time}, the release offset in seconds from the dispatcher's session start.
This offset corresponds to $a_i$ in Section~\ref{sec:benchmark}; setting all offsets to zero produces the burst condition.
Tasks have no deadline fields in the evaluation configuration.

\paragraph{Runtime interpretation.}
The directory tree is a dataset package, not a set of per-task runtime sandboxes.
The merged environment initializes a shared \texttt{/workspace}; task instructions identify the relevant input and output paths within it.
Input files may already exist when the session starts, while the external dispatcher exposes each task only after its release time and allows the agent to inspect all released tasks.
The full schedule and reference solutions are evaluator-side data and are not exposed as agent execution inputs; each task's tests ship with the task and are visible to the agent.
Reference solutions support construction-time validation; after agent execution, each task's \texttt{tests/test.sh} checks the final shared workspace.
Task-level rewards are collected by \texttt{task\_id}, preserving the link between each request and its verified outcome.

\section{Statistical conventions}
\label{app:reproducibility}
\label{app:metric-derivations}

Repeats of a workload are averaged before aggregating across workloads, and
repeated attempts of one workload are not treated as independent samples:
confidence intervals resample workloads, keeping all repeats of a drawn
workload. Rates and ratios aggregate rewards and execution times across
workloads before dividing rather than averaging per-workload ratios.
Individual and stream comparisons use the same scored task set, and every
exclusion is stated with its reason (Appendix~\ref{app:exclusions}).

\begingroup\raggedright
The three L3 examples in Figure~\ref{fig:workload-synthesis} correspond to
\path{tb2-caffe-cifar-10}, \path{tb2-train-fasttext}, and
\path{skillsbench-gravitational-wave-detection}. They illustrate the task pool
rather than asserting that all three co-occur in a particular evaluated
workload.\par
\endgroup

\section{Task exclusions and trial health}
\label{app:additional-results}

\subsection{Excluded tasks}
\label{app:exclusions}

Table~\ref{tab:main-results} excludes eleven tasks from both the individual
and the stream side. A task is excluded when a zero it produced carries no
information about the agent, so counting it as a failure would charge the
agent for a defect in the harness.

\paragraph{Oracle failure.}
\path{skillsbench-video-filler-word-remover}: running the reference solution
and then the grader yields a reward of $0$ (solve script exit code 254). If the
reference solution cannot score the task, an agent scoring $0$ is
uninformative. This task also produced a false zero and is counted once, among
the ten below. Oracle status is measured on the image each workload uses.

\begingroup\raggedright
\paragraph{False zeros (ten tasks).}
A \emph{false zero} is a reward of $0$ written by a grader that never ran. The
test is structural: the grader's output contains no test-result line
(\texttt{\textbackslash d+ (passed|\allowbreak failed|\allowbreak error|\allowbreak skipped)}, or a non-\texttt{pytest}
equivalent), rather than a match on error messages, whose wording varies across
graders. The scan
finds 153 such cells, 101 on the stream side and 52 on the individual side,
concentrated in ten tasks:
\path{tb3-biped-contact-dynamics} (50),
\path{skillsbench-python-scala-translation} (33),
\path{tb3-vllm-deepseek-streaming} (27),
\path{tb3-hof-topology-interpenetration} (19),
\path{tb2-sam-cell-seg} (9),
\path{tb3-telecom-entity-resolution} (8),
\path{tb3-interleaved-vigenere} (3),
\path{skillsbench-energy-market-pricing} (2),
\path{skillsbench-energy-ac-optimal-power-flow} (1), and
\path{skillsbench-video-filler-word-remover} (1).\par
\endgroup

\paragraph{Grader that cannot run (one task).}
\texttt{wildclawbench-04\_\allowbreak Search\_\allowbreak Retrieval\_\allowbreak task\_2\_\allowbreak conflicting\_\allowbreak handling} is
scored by an LLM judge that needs an API key the sandbox does not provide. Every
call fails before judging, and the grader then writes $0$: none of its 189
recorded samples scores above zero, including runs whose transcripts show the
deliverable was produced.

The whole task is dropped rather than only the affected cells. A grader that
could not start in one trial was equally unable to judge its siblings, and
keeping the survivors would compare one arm's luck against another's on the
same broken grader.

\paragraph{Why the oracle does not catch this.}
The defect is that \texttt{environment\_mode = "separate"} is lost during
stream assembly. Upstream, these verifiers run in an isolated environment that
provides \texttt{pytest}; our stream verifier runs roughly twenty graders
inside the container the agent has just used, which provides none. Some
assembled reference solutions install \texttt{pytest} themselves, so the
oracle --- which runs the reference solution and then the
grader --- passes, while an agent, which never runs the reference solution,
fails at \texttt{pytest: command not found}.
\path{tb3-biped-contact-dynamics} is recorded as an oracle pass and still
produced 50 false zeros. An oracle of this shape can only certify that the
reference solution scores; it cannot certify that an agent's work is
scoreable.

\begingroup\raggedright
\paragraph{A grader instability that is not excluded.}
Three tasks --- \path{onebrc-onebrc-400m}, \path{terminalworld-tw_593620} and
\path{terminalworld-tw_528959} --- each grade the same file,
\path{/workspace/result.txt}. Five medium workloads contain two of them.
Graders run once the stream has ended, when the file holds only one task's
output, so in those workloads the two tasks cannot both pass: across all five
systems they never do, in 0 of 75 trials, while each passes in isolation. None
of the exclusion rules above detects this, because the reference solutions
score when each task is checked alone and the graders do run and report an
ordinary failure. We keep the tasks and report this as a limitation of the
graders. It lowers every system's medium stream SR, by at least 3.9 to 6.4
points depending on the system, and leaves the medium ordering unchanged; heavy
workloads contain no such pair.\par
\endgroup

\paragraph{Effect on the reported numbers.}
Excluding these tasks moves medium $\Delta$ by $+0.2$ to $+2.7$ and heavy
$\Delta$ by $-3.8$ to $+5.6$, and it changes the heavy ordering: without the
exclusion Kimi-K3 ($-19.9$) places second behind Claude Opus 5; with it GLM-5.3
($-15.7$) is second and Kimi-K3 ($-19.3$) fourth. Every system still degrades in both tiers
under either convention. The exclusion removes the affected tasks; it does not
recover what the agents would have scored on them, which these runs cannot
measure. Throughput ratios are computed on the same reduced task set as SR.

\subsection{Trial health}
\label{app:health}

A trial is retained when it has a reward record, a transcript, at least eight
trajectory steps, and no output-cap, throttling, unavailable-route or
verifier-download event. Table~\ref{tab:health} applies this rule to every trial
directory of the evaluated medium, heavy and thirty-task jobs, including re-runs
that Table~\ref{tab:main-results} uses only to replace a truncated attempt. The
only failing category is a missing reward record: in those 36 trials the run
stopped before grading. Thirty-four stopped on the harness side --- environment
start-up errors, lost model sessions, or the package-installation failure of
Appendix~\ref{app:codex-setup} --- and two Claude Opus 5 re-runs stopped when the
model refused to continue. None of them is scored.

\begin{table}[htbp]
\caption{Trial health over the evaluated jobs of the five systems.}
\label{tab:health}
\centering
\small
\begin{tabular}{llrr}
\toprule
Setting & Tier & Healthy & No reward record \\
\midrule
Stream & medium & 151 & 3 \\
Stream & heavy & 91 & 2 \\
Stream & thirty-task & 15 & 0 \\
Individual & medium & 223 & 0 \\
Individual & heavy & 634 & 31 \\
Individual & thirty-task & 195 & 0 \\
\midrule
Total & & 1{,}309 & 36 \\
\bottomrule
\end{tabular}
\end{table}

\section{The intake-collapse failure mode}
\label{app:intake-collapse}

On heavy workloads Claude Opus 5 retains 83\% of its isolated success rate while
GLM-5.3 and DeepSeek-V4-Pro retain 74\% and 61\%
(Table~\ref{tab:main-results}). This appendix identifies what the three systems do
differently and rules out the main experimental explanations. All three run the same
harness at the same version, the same reasoning-effort setting, the same
disabled tools, the same dispatcher, the same arrival schedule, the same
container limits and the same scheduling preamble; the deployed per-trial
configurations differ only in the model identifier and the output-token cap
(Appendix~\ref{app:setup-systems}).

\paragraph{The mechanism is how many tasks a run ever holds at once.}
Every one of Claude's 15 trials claimed all 20 released tasks, with a peak
concurrency of 6 to 19. GLM claimed all 20 in 9 trials and between 5 and 10 in
the other six, and in those six its peak concurrency was 1 or 2 --- it executed
the workload one task at a time. DeepSeek did not claim the full workload in
any trial: its peak concurrency is 1 to 3 in 15 of 15, and it claimed 5 to 18
tasks. The resulting never-claimed share is 0\%, 25\% and 48\% of the 300 task
instances per system, and median claim latency, the delay from a task becoming
claimable to being claimed, is 552, 1187 and 6242 seconds.

\paragraph{The collapsed runs are not idle, short, or faulted.}
GLM's six collapsed trials are all \texttt{clean} under both the operational
filter and the looser nearly-clean rule, with zero request-level API errors,
zero output-cap events, zero unavailable-route events and zero verifier
download failures. Four of the six ran past 3.7 of the 4 available hours. They
issue 145 to 281 shell commands and produce 59 to 100~MB of transcript --- four
to six times Claude's median of 15.5~MB on the same workloads --- while calling
the task-intake tool only 12 to 17 times. DeepSeek's median transcript is
102~MB against a median of 14 intake calls. The pattern is sustained effort
concentrated on a handful of claimed tasks, not a failure to act; two of GLM's
collapsed runs also record context-window compaction in their final message,
which makes context pressure a candidate trigger without establishing it.

\paragraph{Intake accounts for part of the gap.}
Splitting GLM's trials on whether they claimed the whole workload separates
part of the deficit:

\begin{center}
\small
\begin{tabular}{lrr}
\toprule
GLM-5.3 heavy trials & $n$ & mean reward \\
\midrule
claimed all 20 tasks & 9 & 0.508 \\
claimed fewer & 6 & 0.348 \\
\midrule
Claude Opus 5, all trials & 15 & 0.625 \\
\bottomrule
\end{tabular}
\end{center}

When GLM claims the whole workload it still scores 11.7 points below Claude on
the same task set; the six collapsed runs account for 6.4 of the 18.1-point
aggregate difference. Both of GLM's heavy trials that exceed the memory
allocation are full-intake runs, so taking on the whole workload carries its
own cost.

\paragraph{For GLM the mode is stochastic; for DeepSeek it is not.}
The same workload yields both regimes across identical repeats: GLM claims 20,
8 and 8 tasks on one heavy workload and 20, 10 and 5 on another, with
byte-identical inputs, arrival schedule and configuration. Single-task-at-a-time
execution is therefore a failure mode a run may enter rather than a fixed
capability level, and Kimi-K3 shows the same instability on the thirty-task
workload, completing 30, 6 and 30 of 30 tasks across three repeats. DeepSeek is
the opposite case: 15 of 15 collapsed, which reads as a policy rather than a
lapse.

\begin{figure}[t]
\centering
\includegraphics[width=\linewidth]{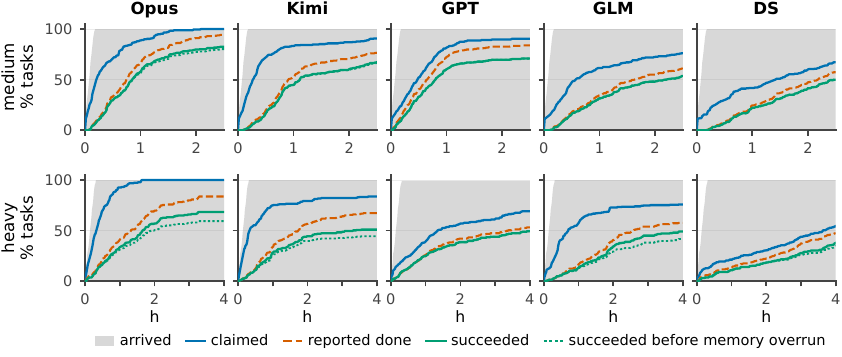}
\caption{Cumulative task progress on the stream clock, as a share of each
trial's scored tasks averaged over trials. Curves show tasks arrived (grey),
claimed (blue), reported done by the agent (orange, dashed), and graded as
successful with reward~1 (green). The dotted green curve removes successes completed
after the trial's memory first exceeded the 8~GiB allocation, as
Table~\ref{tab:main-results} does. Vertical gaps
between consecutive curves are tasks never claimed, tasks claimed but never
reported done, tasks reported done that the grader rejected, and successes
removed by the out-of-memory rule.}
\label{fig:progress}
\end{figure}

\paragraph{Task progress across all five systems.}
Figure~\ref{fig:progress} extends the claim analysis to every system and both
tiers, using the scored task set of Table~\ref{tab:main-results}. Claude Opus 5
claims 89\% of medium and 93\% of heavy tasks within the first hour and every
task by the end of the run, so none of its shortfall comes from intake.
GPT-5.6-sol and DeepSeek-V4-Pro claim gradually rather than up front: after one
hour they hold 39\% and 21\% of heavy tasks, and they finish the run having
claimed 69\% and 55\%. Kimi-K3 shows a third pattern. It claims 83\% of medium
tasks within the first hour and adds fewer than eight points afterwards, while
its reported-done curve keeps rising from 52\% to 77\%, so its remaining work
is claimed but unfinished when the budget ends. At the end of heavy runs the
shares of released tasks never claimed and claimed but not reported done are
0\% and 19.0\% for Claude Opus 5, 25.0\% and 16.3\% for GLM-5.3, 48.3\% and
8.3\% for DeepSeek-V4-Pro, 15.3\% and 17.0\% for Kimi-K3, and 30.3\% and
14.3\% for GPT-5.6-sol. A claimed task that is not reported done can still
earn reward, because the verifier grades the final state, so these shares
overstate the lost reward. Measured in reward instead, with the weights of
Table~\ref{tab:main-results} so that the categories sum to each $\Delta$, tasks never claimed, unfinished at the limit or
completed after a memory overrun cost 7.3 of Claude Opus 5's 13.0 points, 19.4
of GLM-5.3's 15.7, 25.6 of DeepSeek-V4-Pro's 22.7, 16.4 of Kimi-K3's 19.3 and
20.4 of GPT-5.6-sol's 18.1; tasks reported failed cost 0.6 to 4.5 points, and
tasks completed normally cost at most 1.3 points net and in four systems
raise the score, by up to 6.5 points. The gap between the two success
curves appears almost only on heavy, where it is widest for Claude Opus 5
(68\% to 60\%), GLM-5.3 (49\% to 42\%) and Kimi-K3 (51\% to 44\%): work these
systems completed after their memory first exceeded the allocation.

\paragraph{Limits.}
Six of fifteen is a point estimate over five workloads with three repeats; it
does not support a rate. We also do not establish why a run enters the mode ---
the transcripts record the consequence, and the two compaction events are
suggestive, but this collection does not settle the cause. The comparison is
between complete agent systems, so harness-side task-intake ergonomics are part
of what is being measured and cannot be separated from the model here.

\section{Tasks with a zero individual baseline}
\label{app:zero-baseline}

Table~\ref{tab:main-results} keeps tasks on which a system's individual baseline
is zero. They are 4 to 6 of each system's 28--29 scored medium tasks and 6 to 8
of its 46 heavy tasks, between 8\% and 20\% of its stream cells. Because their
individual score is zero by definition, any credit they earn in the stream
makes the system's $\Delta$ less negative.

\paragraph{Most are system-specific, not defects.}
Sixteen tasks have a zero baseline for at least one system. Twelve of them are
solved in isolation by at least one other system, so a zero from three
attempts mostly reflects that system on that task rather than the task. Three
are near zero for every system in both settings
(\path{tb3-satb-audio-transcription}, \path{tb3-pretrain-shard-corruption},
\path{skillsbench-shock-analysis-supply}). One follows a different pattern:
\path{tb2-schemelike-metacircular-eval} scores zero in isolation for all five
systems yet earns 13\% to 40\% in the stream for every one of them, and
\texttt{tb2-overfull-hbox} does the same for four of five. A stream shares one
filesystem across its tasks, so files staged for or produced by another task
are visible to it; we have not established that this is the cause here.

\paragraph{Effect on the reported numbers.}
Table~\ref{tab:zero-baseline} drops each system's zero-baseline tasks from both
of its sides. Every $\Delta$ becomes more negative, by 2.3 to 11.1 points, and
every system still degrades in both tiers. On medium GLM-5.3 and DeepSeek-V4-Pro
swap the last two places; on heavy GLM-5.3 falls from second to fourth, giving
Claude Opus 5, Kimi-K3, GPT-5.6-sol, GLM-5.3, DeepSeek-V4-Pro. We keep the tasks in the headline table
because excluding them per system would give each system a different task set,
and would remove exactly the cases in which shared execution helped.

\begin{table}[htbp]
\caption{Stream-minus-individual difference with and without each system's
zero-baseline tasks. $n$ is zero-baseline tasks over scored tasks; $\Delta$ is
Table~\ref{tab:main-results}; $\Delta_{>0}$ drops the zero-baseline tasks from both
sides.}
\label{tab:zero-baseline}
\centering
\small
\begin{tabular}{lrrrrrr}
\toprule
& \multicolumn{3}{c}{Medium} & \multicolumn{3}{c}{Heavy} \\
\cmidrule(lr){2-4}\cmidrule(l){5-7}
Model & $n$ & $\Delta$ & $\Delta_{>0}$ & $n$ & $\Delta$ & $\Delta_{>0}$ \\
\midrule
Claude Opus 5    & 5/29 & $-$1.8 & $-$5.8 & 6/46 & $-$13.0 & $-$15.3 \\
Kimi-K3          & 4/29 & $-$6.3 & $-$9.6 & 6/46 & $-$19.3 & $-$24.2 \\
GPT-5.6-sol      & 5/28 & $-$14.5 & $-$17.9 & 7/46 & $-$18.1 & $-$25.0 \\
GLM-5.3          & 6/29 & $-$21.1 & $-$30.7 & 8/46 & $-$15.7 & $-$26.8 \\
DeepSeek-V4-Pro  & 6/29 & $-$25.1 & $-$30.0 & 8/46 & $-$22.7 & $-$32.7 \\
\bottomrule
\end{tabular}
\end{table}

\section{Terminal-state accounting}
\label{app:terminal-states}

Two of the six terminal states in Figure~\ref{fig:outcome-decomposition}
depend on which component writes the record, and the comparison in
Appendix~\ref{app:terminal-states-detail} is grouped so that it survives that
dependence.

The fixed policies carry high reported-failed shares --- 19.7\% and 10.7\% of
released tasks on medium and heavy, against 1.7\% and 9.3\% for native
management --- because their driver records a solver session that exits
non-zero as a failed task. An agent-managed run has no such driver and may
simply stop reporting on a task it still holds, which lands in the
claimed-but-unfinished state instead. The two states therefore trade off
against each other for reasons of bookkeeping as much as behavior.

Grouping them resolves the ambiguity: greys are work never served and warm
shades are work served without full credit, regardless of which component wrote the
terminal record. That grouping, not the individual state shares, is what the
policy comparison rests on. The three reported-complete states are unaffected,
since a verifier outcome is written by the same grader in every system, and they
sum to the completion column of Table~\ref{tab:policy-results} in all eight
cells.

\section{Failure-attribution method and adjudication}
\label{app:failure-method}

This appendix records how the failure classification of
Figure~\ref{fig:failure-buckets} is constructed and what the
review-class findings contain. The figure applies the classification to the
trials scored in Table~\ref{tab:main-results}; the review-class scans below cover
all 390 native-stream trials of every tier, including the light tier and the
thirty-task case, so that no trial escapes the manipulation check. The procedure is here because it governs what those counts are
allowed to mean.

\paragraph{Evidence classes.}
We separate three kinds of evidence and never mix them in one number.
\emph{Structured} evidence is stated by an artifact: a harness exit code, a
dispatcher terminal state, a container counter. \emph{Derived} evidence is
computed from a structured series, such as peak memory against the
container's own limit. \emph{Review} evidence is a text match in a
transcript, which locates candidates for inspection and is never reported as
a rate.

Text matching alone cannot support a rate. Generic out-of-memory phrasing
appears 1807 times across 14 of 15 trials in one policy system, while the
collection contains one structurally confirmed occurrence. The matched text is
predominantly a system narrating an error it is reasoning about (``the build
was killed, likely memory, retrying with fewer jobs'') rather than a container
being terminated. Review evidence is therefore reported as a located case list
rather than as a rate.

\paragraph{Resource conflicts.}
Narrow patterns for four conflict classes match in a minority of trials: disk
exhaustion in 56 of 390, process-table or descriptor exhaustion in 47,
file-lock contention in 29, and port collisions in 12. Hit counts exceed trial
counts by two to four times, which indicates individual transcripts repeating
a message rather than many distinct conflicts. A match may be a conflict the
system hit, a conflict it recovered from without consequence, or text it
reproduced while reasoning, and the records do not distinguish these without
inspection. We therefore report a located case list supporting the observation
that shared-container execution produces contention, and derive no conflict
rate.

\paragraph{Adjudicating references to the harness log tree.}
We adjudicated by hand every reference to the harness log tree that our
classifier placed inside an issued command rather than in tool output: 52
candidates in 31 trials, out of 278 references to that tree found in 70 of the
390 trials. No trial references the arrival schedule,
which would be the high-severity case. Most candidates are environment
reconnaissance: directory listings of the harness log tree, which is empty
during the execution phase. Nine are a system rewriting the output paths
inside a task's own test script so that its results land in a scratch
directory, which redirects grader output away from the harness tree rather
than into it. Four are the content of a task's test script appearing in tool
output and misclassified as a command, so the in-command count is itself an
overestimate. The lower-severity pattern, reading a task's own bundled tests,
occurs in 111 trials and is expected: those tests ship with the task.

One candidate is a genuine write. A DeepSeek trial issued
\texttt{echo 1 >} \path{/logs/verifier/reward.txt}, and the harness reward reader
does consult that flat path as its last fallback, after a per-task path. The
write had no effect on the score: the task is recorded at 0.0 in the final
breakdown while the system's own note for it asserts ``reward.txt=1'' and
``3/3 passed'', and the per-task reward record contains 0 with a modification
time after the execution phase ended, so the authoritative grader ran afresh
and contradicted the claim. This is the outcome the fallback ordering
predicts, since the authoritative grader always writes the per-task path that
takes precedence.

The validity argument therefore rests on the contradicted write, which is
empirical, rather than on an argument that the write was impossible: the
agent's log-tree writes do reach the host-side trial directory, and one flat
grader output file carries a modification time inside the execution window.
The reported numbers are unaffected, and the flat fallback path remains
reachable by an agent command.

\section{Details of the experiment setup}
\label{app:setup-details}

All values in this section describe the deployed per-trial configuration and
the recorded trial artifacts. Harbor redacts secrets when
serializing the agent environment, so a masked value indicates a variable
that was set with an unrecoverable value; only an absent key is a defect. We
name environment variables without values.

\subsection{Agent systems as configured}
\label{app:setup-systems}

Within a system, harness version and model identifier are identical across all
jobs and tiers. Claude Opus 5, GLM-5.3, and DeepSeek-V4-Pro run on Claude
Code 2.1.241; GPT-5.6-sol on Codex 0.50.0; Kimi-K3 on Kimi Code 0.36.1. The
Claude Code systems set permission mode to bypass, reasoning effort to
\texttt{high}, and disable the \texttt{WebSearch} and \texttt{WebFetch}
tools. Codex sets reasoning effort to \texttt{high} and uses the
\texttt{responses} wire protocol.

The systems are not identical apart from the model. The Kimi Code system carries
only a version in its harness arguments: no reasoning-effort setting, no
permission mode, and \emph{no tool disabling}. Codex likewise sets no tool
disabling. Web search and fetch are therefore unavailable to three systems and
available to two, which is a configuration difference between the evaluated
systems and not a property of the models. Cross-system comparisons should carry
this caveat.

Requests route through one gateway host. The Claude Code systems set an
Anthropic-compatible base URL and an authentication-token variable; Codex
sets an OpenAI-compatible base URL and key variable. The harness-written
Codex configuration shows that 22 Codex stream trials --- 21 of 27 on medium
and 1 of 15 on heavy --- and 43 of the 91 heavy individual trials reached the
model through a direct route rather than the shared gateway, while the wire protocol stayed
\texttt{responses} throughout. This is a service-level difference in the
collected data.

The output-token cap is set per system through the Claude Code environment. The
GLM systems request 131{,}072 tokens, exactly the gateway ceiling; one
gateway-side token more returns an error. The DeepSeek systems request 393{,}216
tokens, three times that ceiling, so the effective cap is not recoverable
from disk and is recorded as unknown. The Opus systems set no cap variable.

\subsection{Runtime limits}
\label{app:setup-timeouts}

Three independent limits apply, and which one binds differs between the two
settings.

\begin{enumerate}
\item \textbf{Per-request limit}, \texttt{environment.kwargs.request\_timeout}:
nominally the sandbox API request timeout.
\item \textbf{Agent-phase limit}, \texttt{agent.override\_timeout\_sec} and
\texttt{max\_timeout\_sec}.
\item \textbf{Verifier-phase limit}: null in every configuration in the
collection, so a harness default applies.
\end{enumerate}

\begin{table}[htbp]
\caption{Configured runtime limits by setting and tier.
The timeout multiplier is 1.0 and the agent-setup multiplier 5.0 everywhere;
the queue wait limit is 3600~s everywhere.
The binding limit is marked in bold.}
\label{tab:setup-timeouts}
\centering
\small
\begin{tabular}{llrrr}
\toprule
Setting & Tier & Per-request (s) & Agent phase (s) & Container TTL \\
\midrule
Stream & Light & \textbf{7200} & 21600 & 12~h \\
Stream & Medium & \textbf{9000} & 28800 & 16~h \\
Stream & Heavy & \textbf{14400} & 43200 & 24~h \\
Stream & Heavy-plus & \textbf{21600} & 57600 & 30~h \\
Individual & All & 7200 & \textbf{5400} & 4~h \\
Stream (10~h serial policy) & Heavy & \textbf{36000} & 43200 & 24~h \\
\bottomrule
\end{tabular}
\end{table}

In stream execution the per-request limit binds and the agent-phase limit is
never reached: all 15 heavy serial-FIFO trials end their agent phase between
14{,}400 and 14{,}405~seconds, matching the 14{,}400-second per-request
limit, while the agent-phase limit of 43{,}200~seconds is three times higher
and untouched. In individual execution the agent-phase limit binds: the
maximum agent-phase duration is exactly 5400~seconds in all four tiers,
against a 7200-second per-request limit. The execution budget referred to as
4~hours in Sections~\ref{sec:results}--\ref{sec:analysis} is therefore the
per-request limit, and the 10-hour serial policy raises that same limit from
14{,}400 to 36{,}000~seconds, leaving the agent-phase limit unchanged.

The verifier limit appears in no configuration file but is observable: 54
archived trials end their verifier phase at exactly 10{,}801~seconds
(180~minutes), all of them the same energy-market-pricing task variant,
across four systems and four archival batches. The limit is adequate rather than
tight --- one trial completed grading at 7613~seconds and scored full credit,
and that is also the maximum verifier duration in the retained set. No
retained trial exceeds 10{,}000~seconds of grading, so the affected runs are
archived and contribute to no reported number.

\subsection{Measured phase durations}
\label{app:setup-phases}

Computed from the four phase timestamp pairs in each trial record, over the
1750 retained trials. Scores are read from the verifier reward record.

\begin{table}[htbp]
\caption{Phase durations in seconds over all retained trials, and the
agent-setup phase by harness.}
\label{tab:setup-phases}
\centering
\small
\begin{tabular}{lrrr}
\toprule
Phase & Median & p90 & Max \\
\midrule
Environment setup & 201 & 366 & 1076 \\
Agent setup & 13 & 75 & 589 \\
Agent execution & 2403 & 9008 & 21623 \\
Verifier & 33 & 793 & 7613 \\
\midrule
\multicolumn{4}{@{}l}{\emph{Agent setup by harness}} \\
Claude Code & 10 & 43 & 120 \\
Kimi Code & 14 & 57 & 589 \\
Codex & 71 & 110 & 220 \\
\bottomrule
\end{tabular}
\end{table}

Environment setup costs about 200~seconds per trial and is paid once per
container. Individual execution allocates one container per task, so this
fixed cost scales with the task count there while stream execution amortizes
it across the tasks of a workload. This favors stream execution in the
throughput comparison and is one reason those ratios are not efficiency
measurements. Codex agent setup is about seven times slower than Claude
Code's because it is the only harness that installs Node.js, which is also
the origin of the failures in Appendix~\ref{app:codex-setup}.

\subsection{Container and host environment}
\label{app:setup-resources}

Each container is provisioned with four CPUs and 8~GiB of memory, confirmed
in the container control-group telemetry as a quota of 400{,}000~$\mu$s per
100{,}000~$\mu$s period and a memory limit of 8~GiB. The memory limit is
recorded but not enforced: sampled memory can exceed it
(Figure~\ref{fig:resource}). No trial in the
collection records any other quota, so resource headroom is held fixed
throughout and is not a manipulated variable anywhere in this paper.

The sandbox host reports \texttt{x86\_64} on a gVisor kernel. One workload
image configures its package manager against an \texttt{ubuntu-ports} mirror
and fetches arm64 package indices; that is a property of the image's package
configuration and not of the execution architecture. Conflating the two
misdiagnoses the failure in Appendix~\ref{app:codex-setup}. Transcripts are
returned only when a trial ends, because the sandbox is not mounted during
execution, so an absent transcript mid-run is expected and is not evidence of
a stalled run.

\subsection{Task prompt}
\label{app:setup-prompt}

The stream workload prompt is a single 2095-byte text, identical by hash
across all five systems, all tiers, and the guidance, burst, and
resource-labeled conditions. The first trial step carries it for the Claude
Code and Kimi Code systems; for Codex the first step is the harness environment
context and the prompt is the second step.

\begin{quote}\small\ttfamily\raggedright
\# Mission\\[2pt]
You are a CLI agent running as the natural-language interface to a sandbox
container. User work arrives over time as a **stream of tasks** with no fixed
count and no preview of what's coming. Your job is to **complete as many
tasks as possible, as fast as possible**, inside a single wall-clock
budget.\\[4pt]
\# Resources\\[2pt]
This container is provisioned with: **4 CPU cores $\cdot$ 8 GB memory**.
Anything you choose to do --- serial work, parallel work, heavy compute,
light I/O --- shares this single budget. Treat it as the only hardware
constraint that matters.
\end{quote}

The allocation is stated literally in the prompt, so any system that changes the
container quota must change this text as well or the system will plan against
the wrong figure. The guidance condition additionally injects one skill file
of 134 lines through a skills directory variable; every other system leaves that
variable empty.

Individual execution uses no preamble at all. Its first step is the raw task
brief, with no task-pool tools and no instruction about throughput or
parallelism. The individual baseline therefore differs from stream execution
in prompt as well as in isolation, which limits reading the difference
between the two settings as an effect of scheduling alone.

\subsection{Sampling plan}
\label{app:setup-sampling}

Tier sizes were verified against the arrival declarations rather than
assumed: 5, 10, 20, and 30 tasks per workload for light, medium, heavy, and heavy-plus, over 10, 10, 5, and 1 workloads. The last arrival falls between
138 and 358~seconds for light workloads and at 2074~seconds for the single heavy-plus workload. Three attempts are targeted per workload and per
individual task throughout. Trial concurrency varies from 5 to 30 across jobs
as a throughput setting and is not an experimental variable. Planned trial
counts follow: 30 per stream cell at light and medium, 15 at heavy, 3 at heavy-plus, and 69, 42, 93, and 39 per individual cell by tier.

\subsection{GPT-5.6-sol agent-setup failures}
\label{app:codex-setup}

Two gaps in Table~\ref{tab:main-results} belong to the GPT-5.6-sol system: its
medium stream cell holds 27 of 30 planned trials, and its individual baseline
covers 31 of 32 medium task occurrences. Both trace to a single
image-dependent failure in the agent-setup phase, before any model call.

The Codex harness provisions each container with
\texttt{apt-get install -y ... nodejs npm ...}. On one workload image this
command exits 100 with \texttt{E: Unable to correct problems, you have held
broken packages}. The image's apt configuration points at an
\texttt{ubuntu-ports} mirror and fetches arm64 package indices, whereas the
NodeSource repository serves amd64 packages, and the resulting dependency
conflict has no in-image resolution. The three missing stream trials are the
three attempts at that one workload, and the missing individual variant is a
task drawn from the same image.

Three observations fix this as an infrastructure fault rather than a model or
configuration outcome. First, it is deterministic: across 6 live and 23
archived attempts, all 29 trials on this image terminated in setup with no
reward record. Second, it is harness-specific: on the same image the four
Claude Code and Kimi Code systems completed 3 of 3 attempts each, because only
the Codex harness installs Node.js. Third, it is confined to this image; the
same system completed every other planned trial in the tier.

We report the affected cells as incomplete rather than substituting a value.
The stream cell aggregates the 27 trials that ran, and the coverage gap in the
individual baseline and its throughput ratio is recorded in
Appendix~\ref{app:setup-details}. Repairing this requires changing the image's
package sources or the harness Node.js installation path, which alters the
evaluated configuration and is therefore deferred rather than patched into
this evaluation. The container host is x86\_64 on a gVisor kernel; the
arm64 declaration is a property of the image's package configuration, not of
the execution architecture.

\section{Trajectory evidence for the intake account}
\label{app:trajectory-detail}

The dispatcher terminal states attribute delivery differences to task intake.
The recorded trajectories provide an independent view because the tool calls a
system issues show how it spent its turns.
Table~\ref{tab:trajectory-stats} summarizes activity volume and
Table~\ref{tab:tool-mix} its composition.

\begin{table}[htbp]
\caption{Trajectory statistics over the 1750 retained trials, aggregated per system.
Steps are the normalized cross-harness step list; a turn is an agent step carrying at least one tool call, that is one model round trip.
Tool errors count steps whose tool result is flagged as an error.
Token counts are per-trial averages from the trajectory totals; input includes cached prefix tokens.
Sub-agent spawns count delegation tool calls per trial; the Codex harness exposes no delegation tool, so its column is unavailable rather than zero.
Hidden reasoning is reported in three mutually incompatible forms and is \emph{not} comparable across systems: Claude Code emits per-step thinking tokens, Codex emits a per-trial reasoning-token total, Kimi Code emits a count of reasoning steps, and for GLM-5.3 and DeepSeek-V4-Pro the gateway returns a zero thinking-token field (Appendix~\ref{app:trajectory-detail}).
The unit is given in each cell, and dashes mark quantities that are unrecorded or recorded only as zero.}
\label{tab:trajectory-stats}
\centering
\scriptsize
\setlength{\tabcolsep}{3pt}
\begin{tabular}{lrrrrrrrrrrr}
\toprule
& \multicolumn{2}{c}{Steps} & \multicolumn{2}{c}{Turns} & Tool err. & \multicolumn{2}{c}{Tokens/trial} & \multicolumn{2}{c}{Sub-agents} & \multicolumn{2}{c}{Hidden reasoning} \\
\cmidrule(lr){2-3}\cmidrule(lr){4-5}\cmidrule(lr){6-6}\cmidrule(lr){7-8}\cmidrule(lr){9-10}\cmidrule(l){11-12}
Model & p50 & p90 & p50 & p90 & p50 & In (M) & Out (k) & p50 & p90 & p50 & p90 \\
\midrule
Claude Opus 5 & 72 & 310 & 37 & 173 & 2 & 5.4 & 63 & 0 & 5 & 13.9k\,tok & 99.3k\,tok \\
Kimi-K3 & 88 & 300 & 53 & 187 & 3 & 6.4 & 56 & 0 & 6 & 34\,steps & 120\,steps \\
GPT-5.6-sol & 40 & 200 & 28 & 143 & -- & 4.3 & 22 & -- & -- & 5.1k\,tok & 19.1k\,tok \\
GLM-5.3 & 320 & 912 & 161 & 432 & 8 & 15.7 & 280 & 0 & 7 & -- & -- \\
DeepSeek-V4-Pro & 144 & 413 & 75 & 208 & 3 & 7.4 & 161 & 0 & 0 & -- & -- \\
\bottomrule
\end{tabular}
\end{table}

\begin{table}[htbp]
\caption{Tool-call composition per system, as a percentage of all tool calls in the retained trials.
Raw tool names are grouped into families; the task-pool operations are kept separate because they are the actions through which a system takes on and reports work.
\texttt{next} requests the next available task, \texttt{update} reports progress on a held task, and \texttt{spawn} delegates to a sub-agent.
Rows do not sum to 100\% because harness-specific families below 2\% are omitted.}
\label{tab:tool-mix}
\centering
\footnotesize
\begin{tabular}{lrrrrr}
\toprule
Tool family & Opus 5 & Kimi-K3 & GPT-5.6-sol & GLM-5.3 & DeepSeek \\
\midrule
Shell & 65.6 & 51.9 & 77.5 & 73.5 & 78.5 \\
Write/edit & 10.1 & 8.4 & 0.0 & 11.0 & 5.9 \\
Read & 4.6 & 8.5 & 0.0 & 3.9 & 6.9 \\
\midrule
Pool: \texttt{next} & 11.2 & 16.6 & 4.2 & 3.1 & 2.5 \\
Pool: \texttt{update} & 3.7 & 2.3 & 3.7 & 1.2 & 1.5 \\
Pool: \texttt{spawn} & 2.1 & 2.1 & -- & 0.7 & 0.2 \\
\bottomrule
\end{tabular}
\end{table}

\paragraph{Activity volume does not predict delivery.}
GLM-5.3 issues the most activity of any system by a wide margin: a median of 320
steps and 161 model turns per trial against 72 and 37 for Opus, with 15.7M
input and 280k output tokens per trial against Opus's 5.4M and 63k.
It nonetheless delivers lower SR than Opus in every tier.
GPT-5.6-sol occupies the opposite corner, with the fewest steps (median 40)
and the fewest tokens (4.3M input), while placing second on stream SR in both
tiers.
Spending more turns is therefore not the mechanism by which a system delivers
more of a workload, and step or token counts are not proxies for workload
performance.

\paragraph{The tool mix separates the systems along the intake axis.}
Requests for the next available task account for 11.2\% of Opus's tool calls
and 16.6\% of Kimi's, against 3.1\% for GLM and 2.5\% for DeepSeek.
The ordering matches the never-claimed shares of Figure~\ref{fig:mechanism}(c)
(0\% and 12.2\% against 23.8\% and 38.7\%) and is visible per trial: on the
30-task workload Opus issues 69 task requests and 20 delegations while
DeepSeek issues 3 requests and no delegations, spending 84\% of its calls on
the shell.
GLM and DeepSeek thus direct the large activity volumes into repeated work on
tasks they already hold, rather than into acquiring more of the released
workload.

\paragraph{Tool-error counts scale with activity.}
The median trial records 2 failed tool results for Opus and 8 for GLM
(90th percentiles 6 and 24), about four times as many, in proportion to GLM's step count (median 72 vs 320
steps); per step the two error rates are of the same order, about 2--3\%.
The Codex trajectories carry no error flag at all, so its column is
unavailable rather than zero.

\paragraph{Hidden reasoning is recorded in three incompatible units.}
Three systems report hidden reasoning and no two report the same quantity.
Claude Opus 5 carries per-step thinking-token counts (median 13.9k tokens per
trial); GPT-5.6-sol carries a per-trial reasoning-token total (median 5.1k);
Kimi-K3 carries a count of reasoning steps (median 34), not convertible to
tokens. GLM-5.3 and DeepSeek-V4-Pro run on Claude Code but their gateways
return a zero thinking-token field while their trajectories contain reasoning
text (66{,}875 steps for GLM and 24{,}740 for DeepSeek).
A zero or dash in that column indicates what the gateway reported, not how
much the system reasoned; no reasoning-effort ordering can be drawn from this
data.

\paragraph{Sub-agent delegation is a minority behavior.}
The median trial in every system spawns no sub-agent. The 90th percentile is 5
for Opus, 6 for Kimi, and 7 for GLM, against 0 for DeepSeek; GPT-5.6-sol is a
dash because the Codex harness exposes no delegation tool at all, which
differs from DeepSeek's choice not to use one.

\section{Terminal states and resource utilization}
\label{app:terminal-states-detail}

\begin{figure}[htbp]
\centering
\begin{minipage}[t]{0.49\linewidth}
\centering
\includegraphics[width=\linewidth]{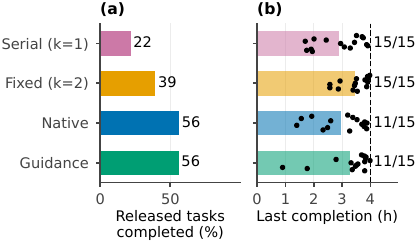}
\caption{Budget use by the four GLM-5.3 policies on heavy workloads, 15 trials each.
(a) Share of released tasks reported completed within the budget; (b) time of each trial's last completion (dots; bars are means), with the number of trials that ran to the 4-hour limit (dashed) at right.
Serial and fixed $k=2$ run to the limit in every trial yet complete 22\% and 39\% of released tasks; native management and guidance complete 56\%, while a quarter and a sixth of their released tasks are never claimed.}
\label{fig:budget}
\end{minipage}\hfill
\begin{minipage}[t]{0.49\linewidth}
\centering
\includegraphics[width=\linewidth]{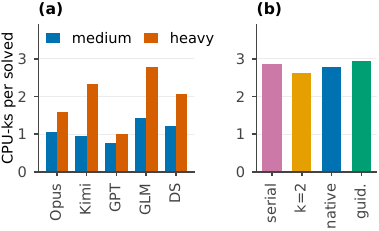}
\caption{Thousands of CPU-seconds per fully solved scored task (setup, grading, and remote inference excluded).
(a) By system and tier under stream execution; (b) GLM-5.3 heavy-tier policies.}
\label{fig:cpu-cost}
\end{minipage}
\end{figure}

\paragraph{The serial deficit is carried by never-claimed work.}
Figure~\ref{fig:outcome-decomposition} counts the scored tasks of
Table~\ref{tab:main-results}. Never-claimed shares fall from 48.3\% to
18.8\% on medium and from 70.0\% to 13.9\% on heavy as the policy moves from
serial to guidance, while the verified share rises from 34.9\% to 55.9\% and
from 18.6\% to 36.7\%. Serial has no unfinished claims, because with $k=1$ a
task is either served to completion or never started.
On heavy, serial and fixed $k=2$ run to the 4-hour limit in every trial
(Figure~\ref{fig:budget}), so their lower delivery is not explained by stopping early.
The serial-minus-native contrast in verified share is $-$0.172
[$-$0.269, $-$0.076] on medium and $-$0.219 [$-$0.329, $-$0.139] on
heavy (percentile bootstrap over workloads), both excluding zero.

\paragraph{Native management's extra held tasks do not raise CPU use.}
Figure~\ref{fig:held-calibration}(a) shows that native management on heavy
holds nearly four tasks at a time and reaches the same CPU utilization as
fixed $k=2$ holding two (27.2\% vs 26.7\%); on medium it holds 1.40 tasks at
the same 14.6\% CPU that serial reaches holding one.
Going from one held task to two does raise it: fixed $k=2$ uses 1.8 and 1.6 times serial's CPU on medium and heavy (Table~\ref{tab:policy-results}).
Every policy leaves at least 70\% of the four provisioned cores idle while
tasks wait.

\paragraph{Self-assessment does not improve with management.}
Of the tasks each policy reported complete, the share the verifier confirmed
is 90.6\%/80.4\% for serial, 88.4\%/87.4\% for fixed $k=2$, 85.5\%/78.7\% for
native management, and 83.4\%/76.8\% for guidance (medium/heavy).
The two agent-managed policies are the least accurate in both tiers, guidance
most of all.
Between 9.4\% and 23.2\% of reported-complete tasks did not verify, so a
self-reported completion is not interchangeable with a verified one.

\paragraph{Compute cost per delivered task.}
CPU-seconds of the agent-execution phase per fully verified task is lowest for
GPT-5.6-sol (759--1020~CPU-s) and highest for GLM (1442--2794).
Across the GLM execution policies, serial FIFO costs more per solved task than
native management in both tiers (1929 vs 1442~CPU-s on medium,
2852 vs 2794 on heavy), because its solved count falls faster than its
compute does. Serial execution is therefore not a compute-saving configuration.

\paragraph{Confirmed memory terminations are rare.}
No native-stream trial carries structured evidence of an out-of-memory
termination, and exactly three trials elsewhere in the collection do (two
GLM-5.3 heavy stream runs under burst arrivals and guidance, and one individual
run, each terminated by signal with harness exit code 137).
The containers expose no usable out-of-memory counters, so a zero reading is
not evidence of absence.

\begin{figure}[htbp]
\centering
\includegraphics[width=\linewidth]{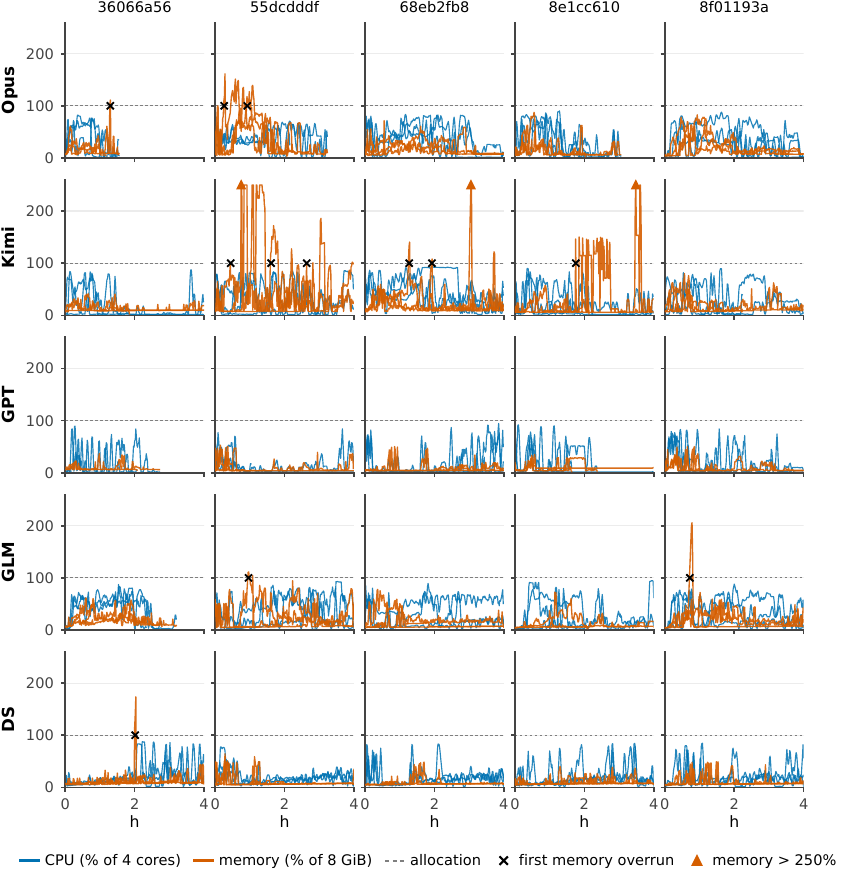}
\caption{CPU and memory utilization of every heavy stream trial, as a share of
the container's allocation (CPU of 4 cores, memory of 8~GiB); one row per
system, one column per heavy workload, three trials per panel. The dashed line
is the allocation; a cross marks a trial's first memory sample above it, from
which point completed tasks are not credited (Table~\ref{tab:main-results}).
Triangles mark memory above 250\%. CPU is smoothed over three minutes; memory is
the per-minute maximum.}
\label{fig:resource}
\end{figure}

\paragraph{Memory overruns on heavy workloads.}
Figure~\ref{fig:resource} traces CPU and memory for every heavy stream trial.
Kimi-K3 exceeds the memory allocation in 6 of its 15 trials, against 3 for
Claude Opus 5, 2 for GLM-5.3, 1 for DeepSeek-V4-Pro and none for GPT-5.6-sol.
Three causes can be separated.
\emph{Workload composition.} Six of the twelve overruns occur on one workload
(\texttt{55dcdddf}: Kimi-K3 in all three trials, Claude Opus 5 in two, GLM-5.3 in
one). It contains \path{tb3-pretrain-shard-corruption}, whose solutions need
about 5~GiB even when the task runs alone: every system's individual runs of it
peak at a median of 5.0--6.5~GiB, so almost any concurrent work pushes the
container past 8~GiB.
\emph{One memory-hungry solution.} On that task Kimi-K3 is the outlier: its
individual runs peak at a median of 6.5~GiB and exceed 8~GiB in 3 of 8, while the
other four systems stay near 5~GiB and never exceed it. One of its stream
overruns happens with that task as the only one in flight.
\emph{Not general inefficiency or over-concurrency.} Across all heavy
individual runs Kimi-K3 exceeds 8~GiB in 1.9\% of them, against 1.3\% for Claude
Opus 5 and GPT-5.6-sol and 4.1\% for GLM-5.3. It also holds fewer tasks at once
than Claude Opus 5 or GLM-5.3 (median realized concurrency 4.5 against 5.4 and
5.9), and four of its six overruns occur with six or fewer tasks in flight,
whereas four of the five Claude Opus 5 and GLM-5.3 overruns occur with 13 or
more. It does not leave work running in the background: it launches the fewest
background processes of any system (0.5\% of its shell commands, against 5.3\%
for Claude Opus 5), and with at most one task in flight
its container holds a median of 0.7~GiB. What remains is exposure: Kimi-K3 runs
11 of its 15 heavy trials to the 4-hour limit, against 6 for Claude Opus 5, so
it spends longer with memory-heavy tasks open.

\paragraph{Contention occurs but is not quantifiable; no score was manipulated.}
Narrow patterns for four resource-conflict classes match in a minority of
the 390 native-stream trials of every tier (disk exhaustion in 56 of 390, process or descriptor exhaustion in 47,
file-lock contention in 29, port collisions in 12), but a match may equally
be a conflict recovered from or text narrated while reasoning.
We adjudicated by hand all 52 command-level references to the harness log tree,
in 31 trials, and report zero confirmed instances of score manipulation. One DeepSeek trial
wrote a reward file on an agent-writable fallback path; the authoritative
grader reran and overrode it, recording 0.0 (Appendix~\ref{app:failure-method}).

\section{Related Work}
\label{app:related}

\paragraph{Computer-use agent evaluation.}
Computer-use benchmarks evaluate agents by the state changes they produce in executable environments.
SWE-bench packages software repositories with executable test suites~\citep{swebench}.
Terminal-Bench and TUA-Bench package each task with a container specification and an in-environment verifier~\citep{terminalbench,tuabench}.
CORE-Bench evaluates the reproduction of computational research from code and data~\citep{corebench}, while TerminalWorld reconstructs Docker environments and tests from recorded terminal workflows~\citep{terminalworld}.
On graphical interfaces, WebArena, VisualWebArena, and OSWorld evaluate browser and desktop agents~\citep{webarena,visualwebarena,osworld}.
Their common limitation for our setting is the evaluation unit: they score one objective in one environment instance, whereas \benchname{} composes multiple independently verifiable objectives in a shared container released over time.

\paragraph{Agents for management and sustained operation.}
A growing line of work evaluates agents as operators of persistent social or economic processes.
Vending-Bench, Business Arena, and E-Commerce Bench ask agents to manage inventories, capital, suppliers, and long-horizon economic outcomes~\citep{vendingbench,businessarena,ecommercebench}.
TheAgentCompany embeds agents in a simulated software company requiring communication, code, and professional tasks~\citep{theagentcompany}.
These environments expose important management capabilities, but their managed object is a domain process; \benchname{} instead studies the role in which the managed object is the user's computational workload under shared resource constraints.

\paragraph{Agent runtimes and resource management.}
AIOS factors scheduling, context and memory management, storage, and access control into an agent-oriented kernel~\citep{aios}.
AgentCgroup characterizes bursty OS-resource use in sandboxed coding agents and applies tool-call-granularity cgroups~\citep{agentcgroup}.
HiveMind coordinates agents sharing rate-limited endpoints through admission control and priority queues~\citep{hivemind}.
SpecBox and PASTE reduce sandbox cold-start and reservation costs through speculative prewarming and joint tool-model scheduling~\citep{specbox,paste}.
These systems improve how a known set of executions consumes particular resources; \benchname{} evaluates the higher-level policy that turns arriving user requests into those executions.

\end{document}